\documentclass[10pt]{article} %

\usepackage[accepted]{rlj} %

\usepackage{amssymb}            %
\usepackage{mathtools}          %
\usepackage{mathrsfs}           %
\usepackage{graphicx}           %
\usepackage{subcaption}         %
\usepackage[space]{grffile}     %
\usepackage{url}                %
\usepackage{lipsum}             %

\usepackage{algorithm}
\usepackage{amsfonts}
\usepackage{algpseudocode}
\usepackage{amsmath}
\usepackage{subcaption}

\usepackage{amsmath}
\usepackage{booktabs}
\usepackage{booktabs} 
\usepackage{multirow}  
\usepackage{makecell} 

\newcommand\showchanges{0}

\newcommand{\chl}[1]{%
    \ifnum 1=\showchanges \relax
        {\color{cyan}#1}\else #1%
    \fi
}

\title{Gaussian Process Aggregation for Root-Parallel Monte Carlo Tree Search with Continuous Actions}

\setrunningtitle{Gaussian Process Aggregation for Root-Parallel Monte Carlo Tree Search with Continuous Actions}

\author{
    Junlin Xiao\textsuperscript{\rm 1,\textdagger},
    Victor-Alexandru Darvariu\textsuperscript{\rm 2},
    Bruno Lacerda\textsuperscript{\rm 3,\textdagger},
    Nick Hawes\textsuperscript{\rm 2}}

\emails{122090599@link.cuhk.edu.cn,
    victord@robots.ox.ac.uk,
    bruno@statefulrobotics.com,
    nickh@robots.ox.ac.uk}

\affiliations{
$^{1}$ School of Science and Engineering, The Chinese University of Hong Kong, Shenzhen\\
$^{2}$ Oxford Robotics Institute, Department of Engineering Science, University of Oxford\\
$^{3}$ Stateful Robotics
\par
$^\dagger$ This work was performed while the author was affiliated with the Oxford Robotics Institute,\\Department of Engineering Science, University of Oxford.
}

\contribution{
    This paper presents Gaussian Process Regression for Root Parallel MCTS (GPR2P). The proposed algorithm builds a principled probabilistic model of the return over the entire action space using information from all the search trees constructed in parallel. Inference is performed using this model to inform the selection of the final aggregated action. This results in promising empirical performance with a modest increase in inference time.
    }
    {
    Given the serial nature of MCTS, root-parallel approaches are often used in practice when wall clock time is limited. In domains with continuous actions, the typical majority voting strategy~\citep{soejima2010evaluating} used in domains with a finite number of discrete actions is not applicable. The state-of-the-art in this area is the work of~\citet{kurzer2020parallelization}, who propose two aggregation methods that reuse information across related actions. However, the aggregation of information between related actions is performed in an ad-hoc manner, and the methods \chl{only select among actions that were tried in the environment}. Our method addresses both these shortcomings.   
    }

\contribution{
    This paper presents a comprehensive empirical evaluation of GPR2P against prior approaches across six benchmark environments. The results demonstrate consistent performance improvements. To the best of our knowledge, this study also constitutes the most extensive evaluation to date of aggregation strategies for root-parallel MCTS in continuous action domains.
    }
    {
    Despite the practical relevance of root-parallel MCTS, aggregation methods have not been extensively benchmarked in domains with continuous actions. The work of~\citet{kurzer2020parallelization} only considers a single environment and compares the proposed root-parallel methods with leaf parallelization and single-thread MCTS, omitting other root-parallel aggregation baselines. In contrast, we study the performance of root-parallel aggregation methods in six environments and compare with a wide range of baselines.        
    }


\summary{Monte Carlo Tree Search is a cornerstone algorithm for online planning, and its root-parallel variant is widely used when wall clock time is limited but best performance is desired. In environments with continuous action spaces, how to best aggregate statistics from different threads is an important yet underexplored question. In this work, we introduce a method that uses Gaussian Process Regression to obtain value estimates for promising actions that were not trialed in the environment. We perform a systematic evaluation across 6 different domains, demonstrating that our approach outperforms existing aggregation strategies while requiring a modest increase in inference time.
}

\begin{document}

\makeCover  %
\maketitle  %

\begin{abstract}
Monte Carlo Tree Search is a cornerstone algorithm for online planning, and its root-parallel variant is widely used when wall clock time is limited but best performance is desired. In environments with continuous action spaces, how to best aggregate statistics from different threads is an important yet underexplored question. In this work, we introduce a method that uses Gaussian Process Regression to obtain value estimates for promising actions that were not trialed in the environment. We perform a systematic evaluation across 6 different domains, demonstrating that our approach outperforms existing aggregation strategies while requiring a modest increase in inference time.
\end{abstract}

\section{Introduction}
Monte Carlo Tree Search (MCTS) is a widely used online planning algorithm. Its anytime nature, ability to plan from the present state, and requirement for only sample-based access to the transition and reward functions have lead to successful applications in practical domains with large state and action spaces~\citep{silver2010monte,silver2016mastering}. MCTS relies on the quality of simulation returns, which poses challenges for identifying strong actions when only limited time or simulation budgets are available. Many existing works \citep{sokota2021monte,silver2016mastering, cazenave2015generalized} focus on modifying the original sampling mechanism to improve the performance of single-threaded MCTS. Root-parallel MCTS, however, provides a complementary direction: by aggregating the results of multiple independent MCTS threads, it can naturally leverage state-of-the-art advances developed for individual MCTS variants. This aggregation strategy offers a flexible and powerful framework that allows different MCTS variants to contribute according to their respective strengths. Despite its potential, this direction remains largely underexplored.

Root-parallel MCTS can generate a large amount of data from multiple parallel trees. However, designing an effective aggregation mechanism to choose the final action based on this data remains a crucial problem. This challenge is particularly pronounced in continuous action domains. Unlike in domains with discrete actions, each sampled action is unique, rendering the typical Majority Voting~\citep{soejima2010evaluating} approach inapplicable. Adopting the action with highest return (Max) and visit count (Most Visited) across the threads are sensible baseline strategies. Furthermore, state-of-the-art methods aim to exploit the relationships among the returns of individual actions~\citep{kurzer2020parallelization}. Intuitively, when starting from the same state, similar actions (e.g., as quantified by a distance metric) will drive the agent towards similar subsequent states, which in turn increases the likelihood of obtaining similar simulation returns compared to completely different actions.

However, important limitations remain. Firstly, such methods lack a mechanism for interpolating between sampled actions. While the returns of actions will influence those of nearby ones as quantified by the distance metric, the methods \chl{do not} output actions that were not trialed in the tree, a significant downside in action spaces with meaningful structure. Secondly, the ranking schemes adopted by current approaches are typically fixed and manually designed, which may not generalize consistently across different tasks. Thirdly, the benefit of such approaches relative to single thread MCTS and simple baselines remains unclear. Currently, a thorough evaluation of aggregation strategies for root-parallel MCTS is lacking in the literature. The contributions of our paper therefore aim to address these limitations:
\begin{enumerate}
    \item We propose \textit{Gaussian Process Regression for Root Parallel MCTS (GPR2P)}. Unlike previous approaches that select only sampled actions, we construct a principled statistical model of the return over the entire action space. This capability is especially important when promising actions are difficult to sample and only a limited number of simulations can be performed.
    \item We carry out a comprehensive empirical evaluation comparing GPR2P to previous techniques across six different environments, demonstrating that it yields better performance while requiring a modest increase in inference time. To our knowledge, this also represents the most extensive evaluation of aggregation strategies for root-parallel MCTS in continuous domains to date.
\end{enumerate}

\section{Background}

\begin{figure*}[t]
    \centering
    \includegraphics[width=\textwidth]{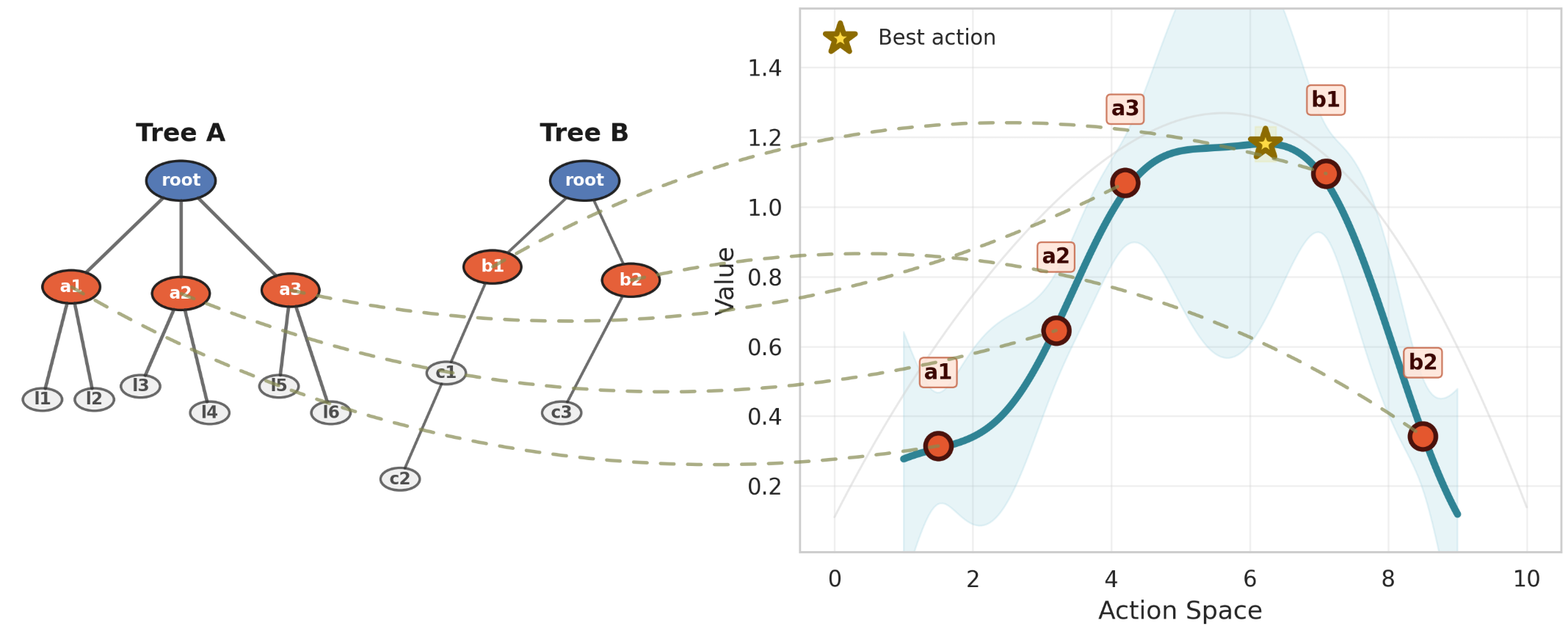}
    \caption{Illustration of the GPR2P method, which uses Gaussian Process Regression to perform aggregation in root-parallel MCTS. Unlike existing methods, \chl{GPR2P estimates the return for and selects actions} that were not sampled in the tree.}
    \label{fig:summary}
\end{figure*}

\textbf{Markov Decision Processes.}
In this work, we consider an agent interacting with an environment under uncertain dynamics. This problem can be described as a Markov Decision Process (MDP) \citep{bellman1957markovian}, which is formally defined by a tuple $\langle S, A, T, R, \gamma \rangle$ where (i) $S \subseteq \mathbb{R}^N$ is the N-dimensional state space; (ii) $A\subseteq \mathbb{R}^D$ is the D-dimensional action space; (iii) $T(s,a,s')$ is the transition function, which represents the probability of reaching $s'$ from $s$ by taking action $a$; (iv) $R$ is the reward function $R(s,a)$ governing the reward received when taking action $a$ in state $s$; (v) $\gamma \in [0,1]$ is the discount factor.
In MDPs, a policy $\pi$ maps states to actions, and its value function $V^\pi (s)$ denotes the expected sum of discounted rewards obtained by starting from state $s$ and following $\pi$. The expected discounted reward from time $t = 0$ to time $t = T$ is calculated as $\sum_{t=0}^T \gamma ^t r_t$, where $r_t$ denotes the reward received at time $t$. There exists at least one optimal policy $\pi ^ *$ which optimizes the value function $V^\pi (s), \forall s \in S$.

\textbf{Monte Carlo Tree Search.}
Monte Carlo Tree Search (MCTS) is a best-first search algorithm \citep{coulom2006efficient,chaslot2008parallel} that performs randomized explorations in the search space and asymptotically converges to the optimal solution, which corresponds to the optimal trajectory in a Markov Decision Process (MDP). The search process is guided by selection strategies designed to balance the trade-off between exploring new actions and exploiting promising ones based on existing samples~\citep{browne2012survey}. Among these strategies, the Upper Confidence Bounds for Trees (UCT) algorithm \citep{kocsis2006bandit} is one of the most widely adopted, and it also serves as the foundation of our method. The UCT selection rule is defined as
$
    UCT(s,a) = Q(s,a) + C\sqrt{\frac{2\ln N(s)}{N(s,a)}}
$
where the first term represents the estimated value of a specific state–action pair. A higher return value for this pair increases its UCT score, thereby raising the probability of it being selected. The second term controls exploration. Here, \chl{$C$ determines the strength of exploration relative to exploitation, and $N(s,a)$ denotes the number of visits for state–action pair $(s,a)$. When $N(s,a)$ is small}, the exploration term becomes large, indicating that action $a$ has been rarely tried for state $s$, thus encouraging exploration. MCTS repeatedly performs the four key steps of selection, expansion, simulation, and backpropagation until the computational budget is exhausted.

\textbf{Progressive Widening.}
In MCTS, the action selection equation requires every possible action to have at least one simulation. It is suitable when the action space is discrete and finite since it allows for exploring every possible action. However, in continuous action spaces, it is impossible to explore every action. Progressive Widening (PW) is designed to address this problem~\citep{chaslot2008progressive, coulom2007computing}. This approach requires a node to be visited sufficiently many times in order to expand the action space. The relationship between the visit count $N(s)$ of a state $s$ and its action space $A(s)$ is described as
$
    \lvert A(s) \rvert = \lfloor c*N(s)^\alpha \rfloor  
$, 
where the parameters $c$ and $\alpha$ control the rate at which the action space of a state is widened. In other words, a new sampled action is added to the tree only when the visit count of $s$ exceeds a threshold. While PW is effective in many continuous domains, it is insufficient when transitions are stochastic, i.e., performing the same action in the same state can lead to different successor states. To address this issue, Double Progressive Widening (DPW) was introduced by~\citet{couetoux2011continuous}. In DPW, increasing the number of recorded successor states requires the corresponding action to be visited sufficiently many times. Formally, the number of successor states associated with action 
$a$ is given by
$
    \lvert \text{Succ}(a) \rvert = \lfloor d*N(s,a)^\beta \rfloor  
$, 
where the parameters $d$ and $\beta$ control the rate at which the successor set of an action is expanded.

\textbf{Parallelization}
Parallelizing MCTS has shown great potential for improving search efficiency \citep{enzenberger2009lock,gelly2008parallelization}. There are three main types of parallelization: leaf parallelization, root parallelization, and tree parallelization \citep{cazenave2007parallelization,chaslot2008parallel}. These approaches are typically distinguished and named according to the position in the search process where parallel computation is applied. Root parallelization constructs multiple independent search trees from the same root state, and each thread performs an independent search under a shared computational budget or time limit. After termination, the trees are aggregated to determine the action to execute by leveraging the returns of actions sampled at the shared root across all threads. Once the selected action is applied and the environment transitions to a new state, the same sampling and aggregation procedure is repeated, making the overall approach an instance of online planning. The effectiveness of this framework largely depends on the aggregation strategy, which determines how information from different threads is combined~\citep{steinmetz2020more}. The parallelization mechanism does not impose any specific requirements on the underlying MCTS sampling procedure. On the contrary, advanced MCTS variants that produce higher-quality sampling statistics can further enhance the merging process by providing more reliable information for aggregation. In this work, we adopt UCT as the underlying MCTS algorithm, as it has been widely used and evaluated in prior related studies. We review existing aggregation methods and introduce our proposed approach in the Methods section.

\textbf{Gaussian Process Regression}
GPR is a non-parametric Bayesian method for modeling the relationship between inputs and outputs \citep{williams1995gaussian}. Unlike parametric models, it assumes no fixed functional form and instead infers the underlying structure directly from data. Its foundation lies in Gaussian Processes (GPs), which extend multivariate Gaussian distributions to infinite-dimensional function spaces \citep{rasmussen2006gaussian}. In practice, the mean function is typically set to zero, making the covariance function $k(x,x')$ the central component that encodes correlations between observations. A commonly used choice is the Radial Basis Function (RBF) kernel \citep{scholkopf1997comparing}:
$
    k(x,x') = \sigma_f^2 \exp\left[-\frac{\|x - x'\|^2}{2l^2}\right] + \sigma_n^2 I_n
    \label{eq:RBF}
$.
Here, $\sigma_f^2$ represents the signal variance and \chl{$\sigma_n^2$ accounts for observation noise.} The kernel function $k(x,x')$ is designed such that points that are close in the input space exhibit higher similarity, while points that are far apart have negligible correlation. The length-scale parameter $l$ controls the rate at which this similarity decays. To perform regression with GPs, we aim to compute the predictive mean $y_{*}$ corresponding to a new input $x_{*}$. Under the GP assumption, the joint distribution of the training outputs and the test output 
$y_{*}$ follows a multivariate Gaussian~\citep{mackay1998introduction}. From the collected data, an approximate output distribution can be derived, which yields the mean, variance, and confidence interval.

\section{Methods}~\label{sec:methods}
In this section, we present several commonly used aggregation methods for root-parallel MCTS, and then introduce our proposed method. As in prior work, the information to be aggregated comprises the estimated value of each state-action pair and its visit count. In root-parallel MCTS, all threads begin from the same root state and independently construct their own search trees. Consequently, when aggregating results, all pairs correspond to the same root state $s_0$.

\subsection{Previous Root-Parallel Aggregation Methods}
\textbf{Max and Most Visited.}
The \textit{Max} algorithm is a straightforward aggregation strategy: it chooses the state–action pair with the highest value across all threads. Concretely:
$
    a_{\text{final}} = \arg\max_{a \in \mathcal{A}_{\text{final}}} Q(s_0,a)
$, 
where $\mathcal{A}_{\text{final}}$ is the set containing all sampled actions from all threads.
The \textit{Most Visited} algorithm selects the action with the highest visit count, as in MCTS good actions have a significantly larger number of simulations:
$
    a_{\text{final}} = \arg\max_{a \in \mathcal{A}_{\text{final}}} N(s_0,a).
$

\textbf{Similarity Vote and Similarity Merge.}
Unlike previous algorithms that evaluate actions independently, \citet{kurzer2020parallelization} proposed two methods that exploit action similarity to aggregate all trees in the \chl{tree} set $\mathcal{X}$. A similarity matrix $K$, derived from Euclidean distances between actions, links their values. The parameter $\phi$ \chl{is used in constructing $K$ and} controls how distances map to similarities: a large $\phi$ yields highly selective similarities, while a small $\phi$ produces broader connections.
\textit{Similarity Vote} (Algorithm~\ref{alg:sim_vote}) extends a root-parallel voting scheme~\citep{soejima2010evaluating} to continuous action spaces. Using the similarity matrix $K$, it updates each action value through a weighted sum of the values of all other actions. 
\textit{Similarity Merge} (Algorithm~\ref{alg:sim_merge}) follows the same principle but additionally incorporates visit counts, reflecting their reliability \citep{chaslot2008parallel}. Actions with higher visit counts therefore contribute more strongly during aggregation. By default, Similarity Vote performs poorly when rewards can take negative values, because its reliability assessment assumes that higher returns indicate better and more consistent actions. When all values are negative, this assumption breaks down. To address this limitation of Similarity Vote, we add a constant positive offset to all returns in environments with negative rewards.

\subsection{GPR2P}

\begin{algorithm}
\caption{GPR2P}
\label{alg:gpr2p}
\begin{algorithmic}[1]

\Require Collection of trees $\mathcal{X}$, action space $\mathcal{A}$, 
kernel parameters $(\sigma_f,l,\sigma_n)$, threshold $\tau$

\State $\mathcal{A}_{\text{valid}} \gets \{ a \in \mathcal{X} \mid N(s_0,a) \ge \tau \}$

\For{$a_i \in \mathcal{A}_{\text{valid}}$}
    \State $y_i \gets Q(s_0,a_i)$
\EndFor

\State $\chl{n \gets |\mathcal{A}_{\text{valid}}|}, \quad X \gets [a_1,\dots,a_n], \quad y \gets [y_1,\dots,y_n]^T$

\For{$i=1$ to $n$}
    \For{$j=1$ to $n$}
        \State $K_{ij} \gets \sigma_f^2 \exp\!\left(-\frac{\|a_i-a_j\|^2}{2l^2}\right)$
    \EndFor
\EndFor

\State $\mu(a) \gets k(a,X)(K+\sigma_n^2 I)^{-1}y$

\State \Return $a^* \gets \arg\max_{a \in \mathcal{A}} \mu(a)$

\end{algorithmic}
\end{algorithm}

While Similarity Vote and Similarity Merge link information from related actions, this is not performed in a statistically principled way. \chl{Furthermore, these methods only select among actions that were tried in the environment, rather than explicitly optimizing over untried actions.} When high-quality actions are difficult to discover, or when computational resources are insufficient to support extensive sampling, these methods struggle to consistently identify good actions. Our proposed \textit{GPR2P} method applies Gaussian Process Regression to estimate action values across the entire action space in a principled manner. Additionally, unlike methods that restrict selection to sampled actions, our method estimates returns over the full action space. 

Pseudocode for GPR2P is given in Algorithm~\ref{alg:gpr2p}. It first applies a visit-count threshold $\tau$: only actions whose visit counts exceed this threshold are retained for regression. This serves as a simple reliability filter that excludes insufficiently explored actions. For the retained actions, GPR is performed on their value estimates. In our implementation, we use the Radial Basis Function (RBF) kernel. The algorithm then selects the optimal action across the whole action space based on the model (line 7). We opt for selecting the action with the highest estimated mean return, as we found it performed best in preliminary experiments. Other selection criteria, such as picking actions with the highest confidence interval upper limit, are also applicable. 

\textbf{Computational Complexity.}
While Gaussian Process Regression incurs a cubic complexity of $O(M^3)$ with respect to the number of data points $M$, the proposed filtering mechanism effectively limits $M$ by retaining only highly visited actions. \chl{In our experiments, the number of data points pairing actions and return estimates that were retained for GPR varied across settings, with a maximum of approximately $400$.} As the number of trials increases, these high-visit actions provide more reliable signals, thereby reducing computational overhead and improving the quality of the regression inputs. As shown in Table~\ref{tab:Time comparison table} in our evaluation, this yields strong empirical performance and good scalability on the considered problems.
For higher-dimensional action spaces, computing the RBF kernel matrix (line 4) incurs a computational cost of $O(M^2 d)$, where $d$ denotes the action dimension. Since $M$ remains bounded due to the filtering strategy, GPR2P would also scale efficiently to higher-dimensional action spaces compared to dense discretization.

\begin{figure*}[t]
\centering
    \begin{subfigure}{0.3\textwidth}
        \includegraphics[width=\linewidth, keepaspectratio]{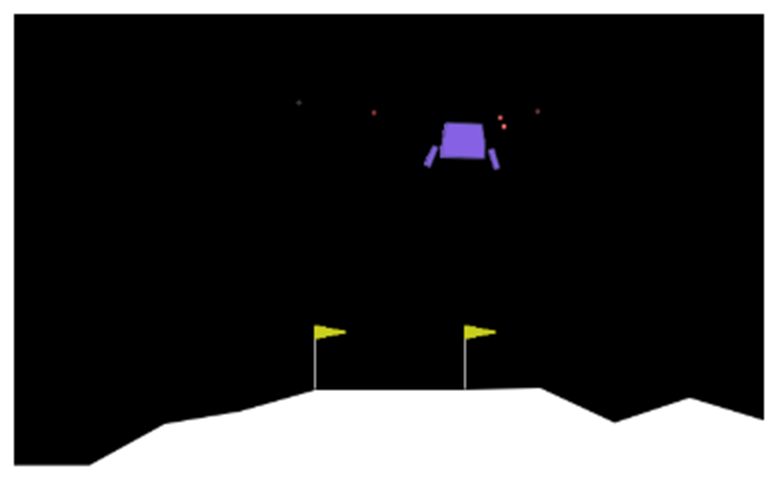}
        \caption{Lunar Lander}
        \label{fig:Lunar Lander}
    \end{subfigure}
    \begin{subfigure}{0.3\textwidth}
        \includegraphics[width=\linewidth, keepaspectratio]{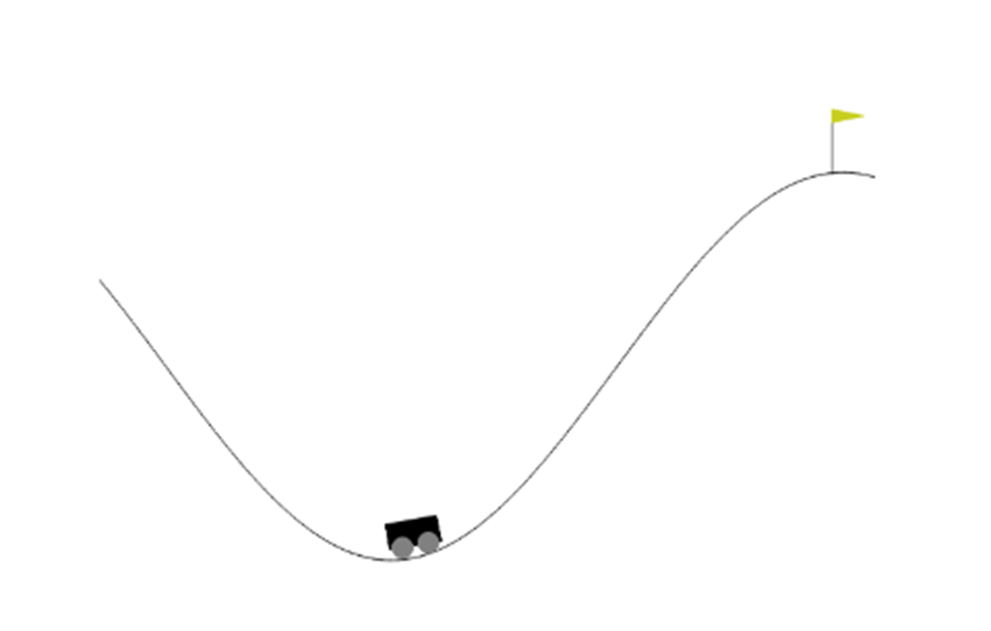}
        \caption{Mountain Car}
        \label{fig:Mountain Car}
    \end{subfigure}
    \begin{subfigure}{0.3\textwidth}
        \centering
        \includegraphics[width=0.8\linewidth, keepaspectratio]{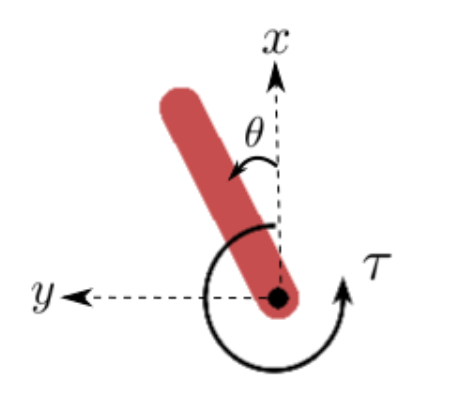}
        \caption{Pendulum}
        \label{fig:Pendulum}
    \end{subfigure}

    \begin{subfigure}{0.3\textwidth}
        \includegraphics[width=\linewidth, keepaspectratio]{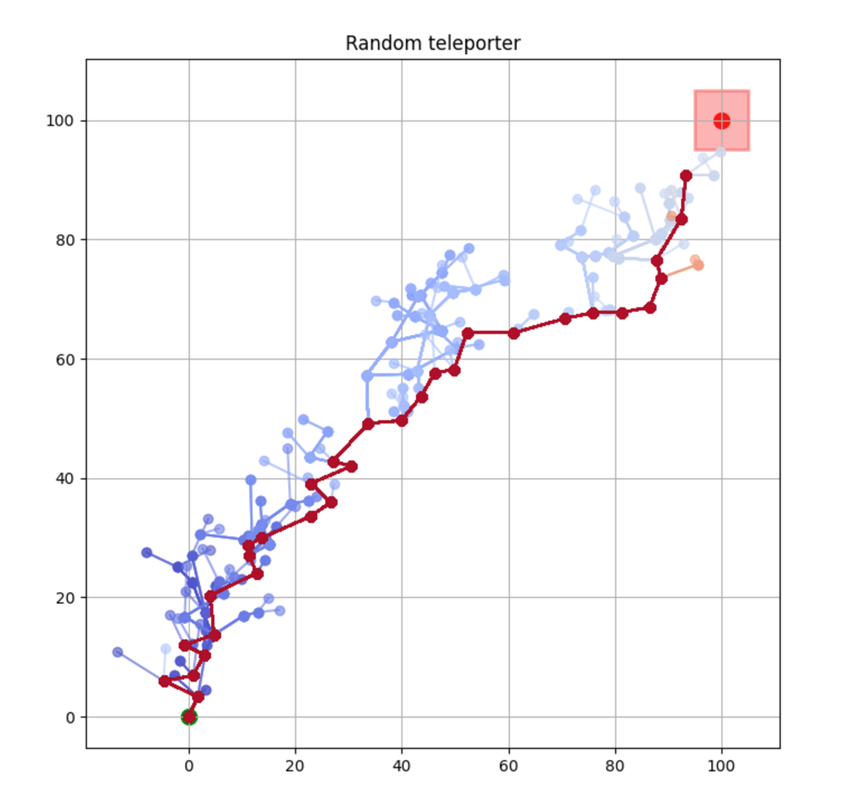}
        \caption{Random Teleporter}
        \label{fig:Random Teleporter}
    \end{subfigure}
    \begin{subfigure}{0.3\textwidth}
        \includegraphics[width=\linewidth, keepaspectratio]{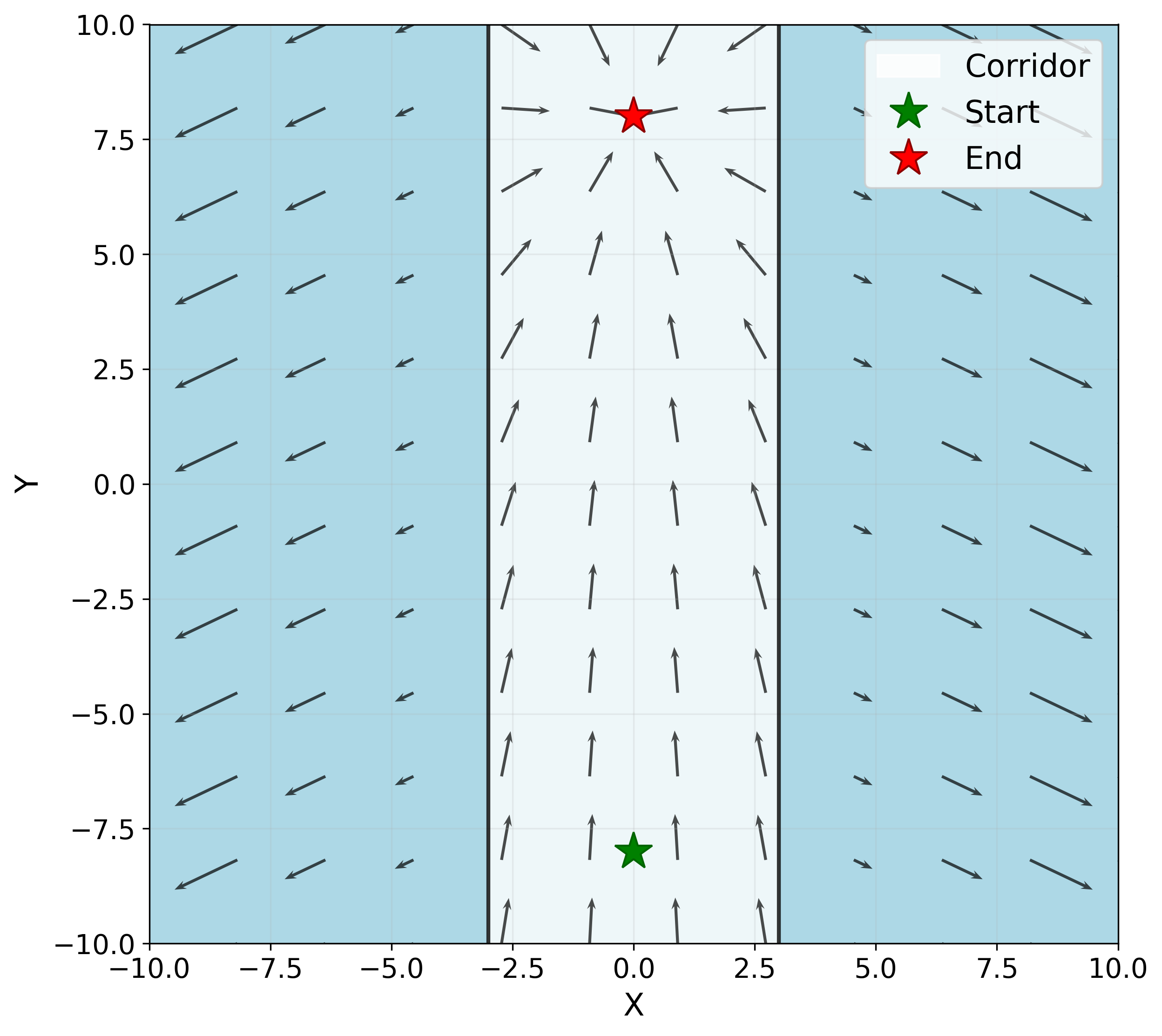}
        \caption{Wide Corridor}
        \label{fig:Wide Corridor}
    \end{subfigure}
    \begin{subfigure}{0.3\textwidth}
        \includegraphics[width=\linewidth, keepaspectratio]{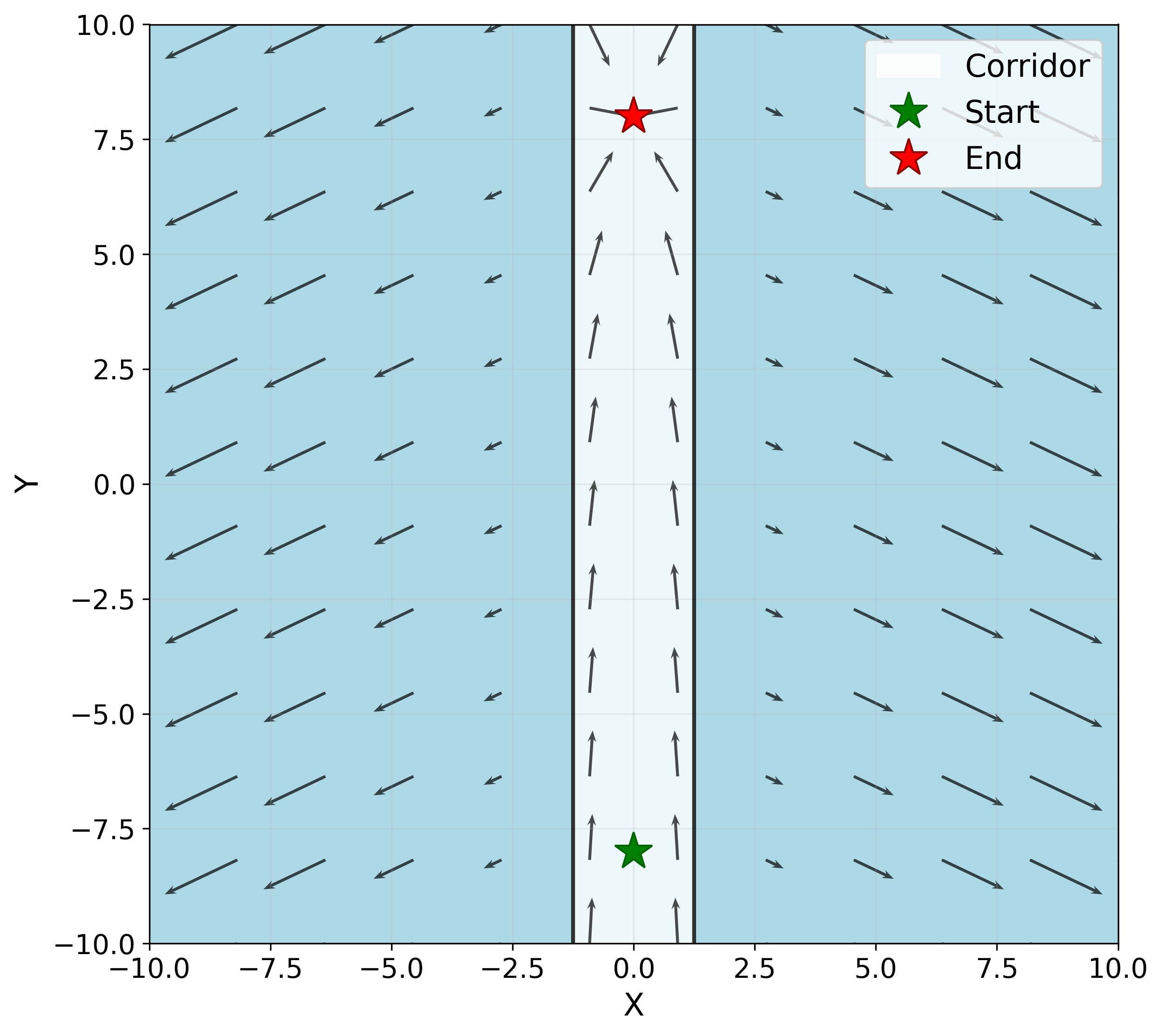}
        \caption{Narrow Corridor}
        \label{fig:Narrow Corridor}
    \end{subfigure}

    \caption{Illustrations of the environments considered in our evaluation. This includes Gymnasium environments (top row) and simplified self-designed environments with stochastic transitions (bottom row). All environments use continuous action spaces.}
    \label{fig:envs}
    
\end{figure*}

\section{Experimental Evaluation}

\begin{figure*}[t]
\centering
    \begin{subfigure}{0.3\textwidth}
        \includegraphics[width=\linewidth, keepaspectratio]{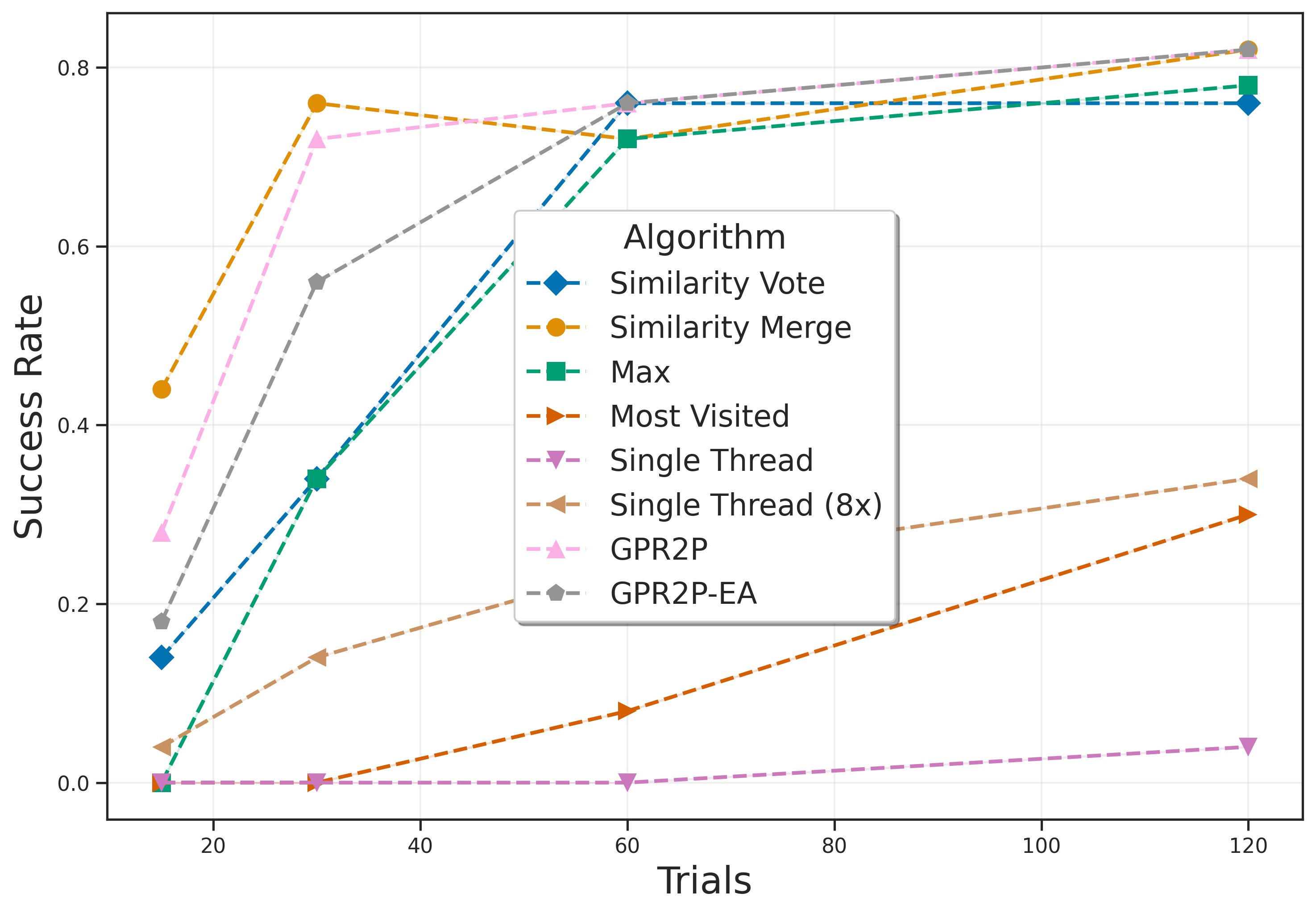}
        \caption{Lunar Lander}
        \label{result:Lunar Lander}
    \end{subfigure}
    \begin{subfigure}{0.3\textwidth}
        \includegraphics[width=\linewidth, keepaspectratio]{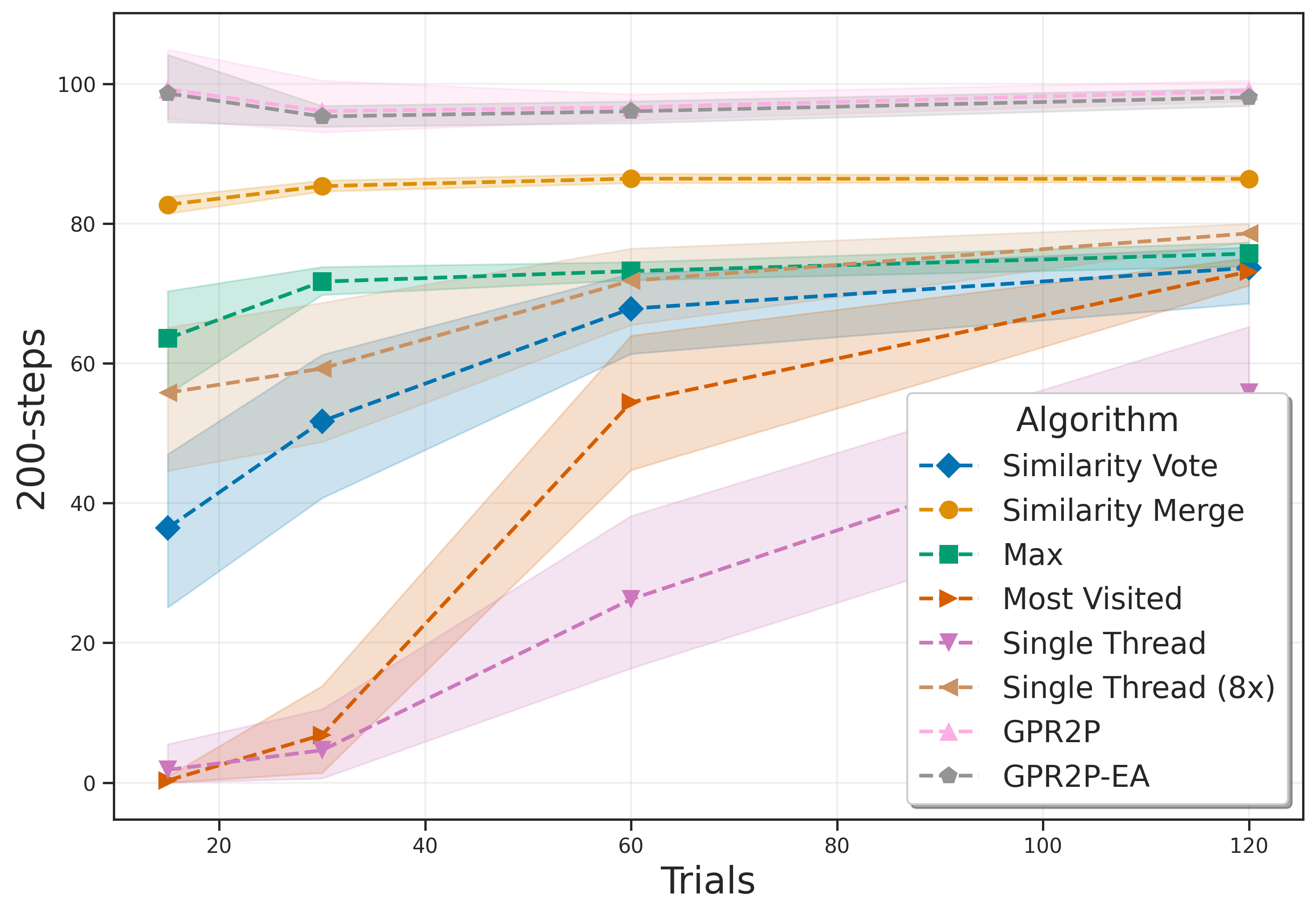}
        \caption{Mountain Car}
        \label{result:Mountain Car}
    \end{subfigure}
    \begin{subfigure}{0.3\textwidth}
        \includegraphics[width=\linewidth, keepaspectratio]{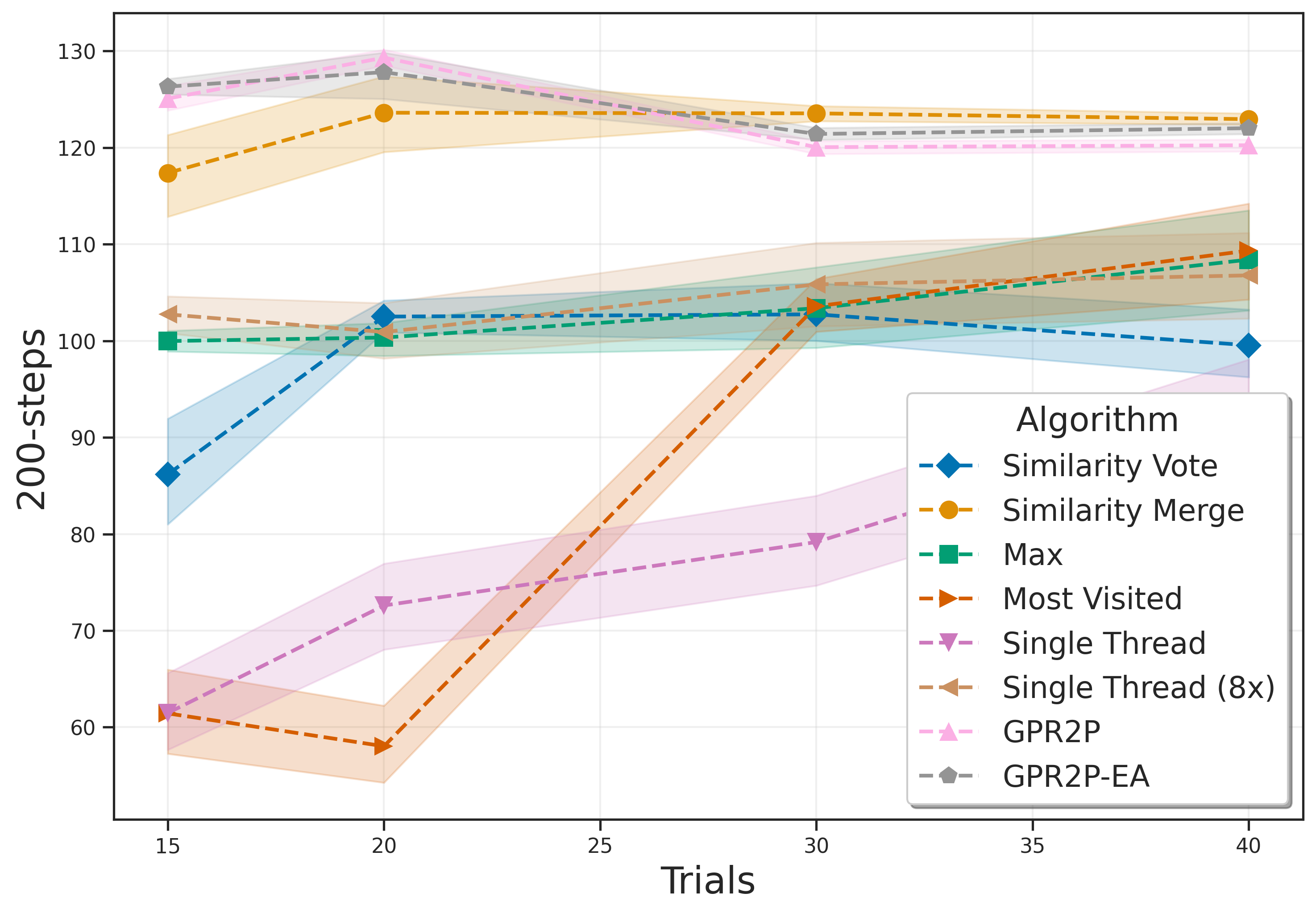}
        \caption{Pendulum}
        \label{result:Pendulum}
    \end{subfigure}

    \begin{subfigure}{0.3\textwidth}
        \includegraphics[width=\linewidth, keepaspectratio]{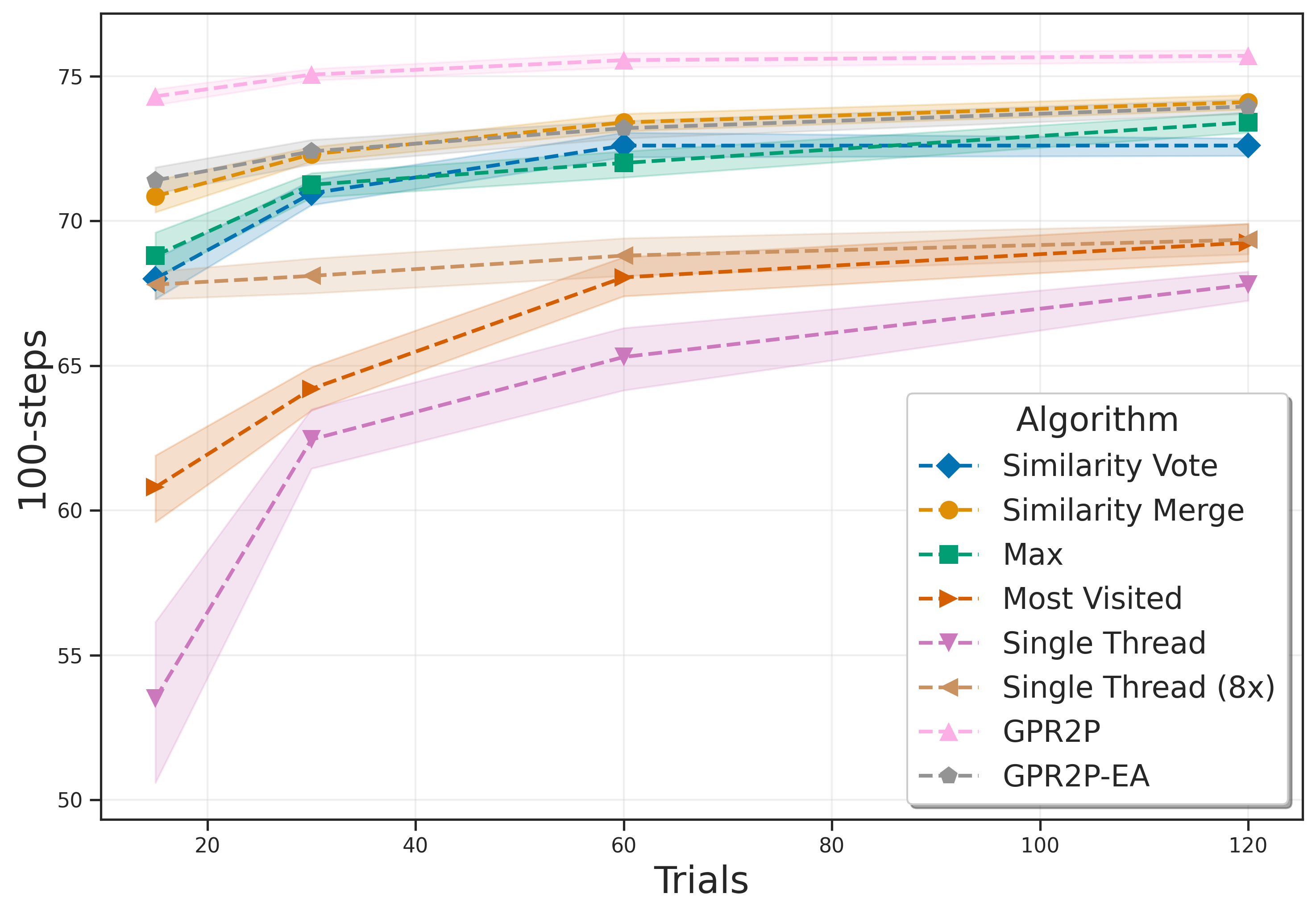}
        \caption{Random Teleporter}
        \label{result:Random Teleporter}
    \end{subfigure}
    \begin{subfigure}{0.3\textwidth}
        \includegraphics[width=\linewidth, keepaspectratio]{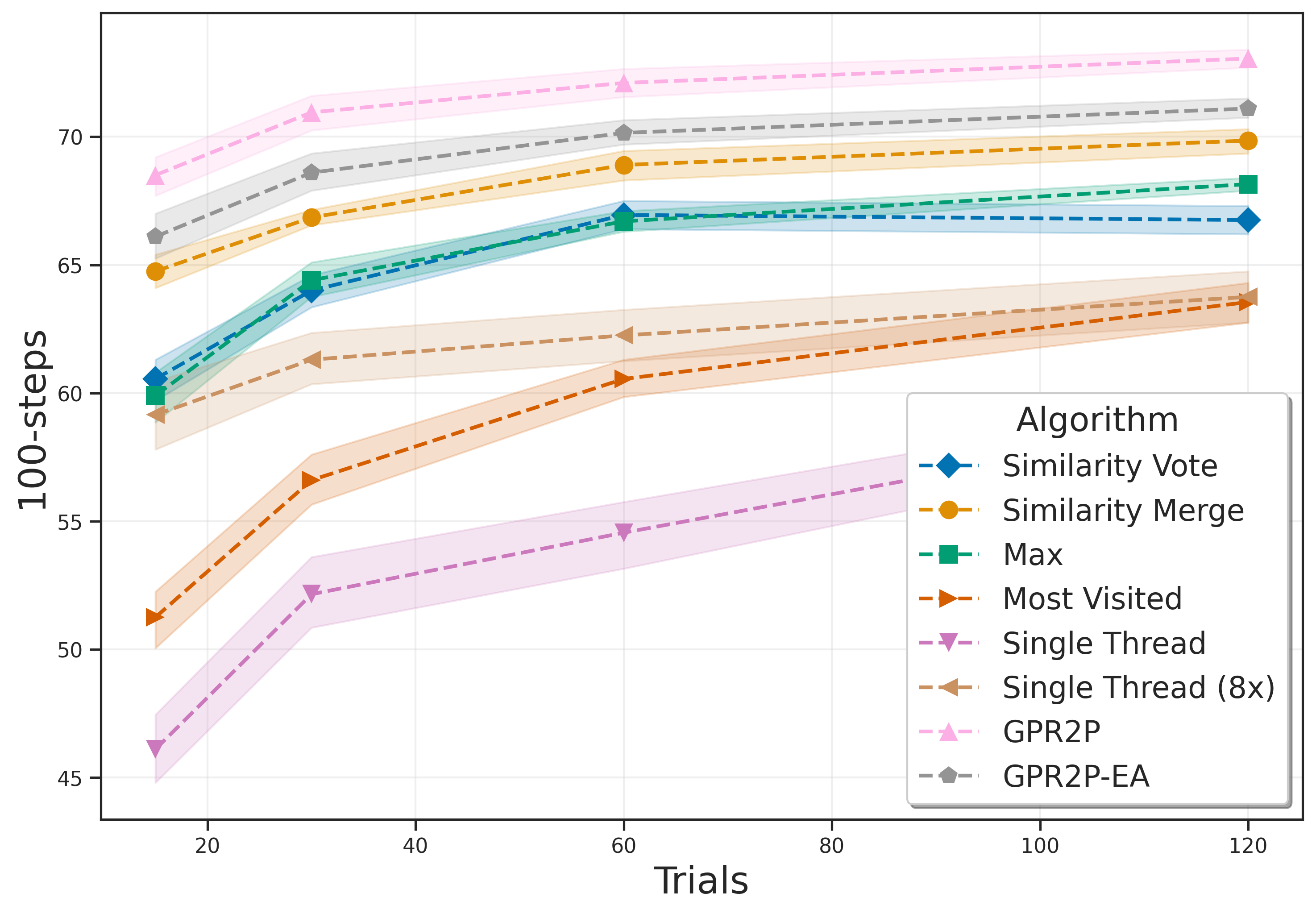}
        \caption{Wide Corridor}
        \label{result:Wide Corridor}
    \end{subfigure}
    \begin{subfigure}{0.3\textwidth}
        \includegraphics[width=\linewidth, keepaspectratio]{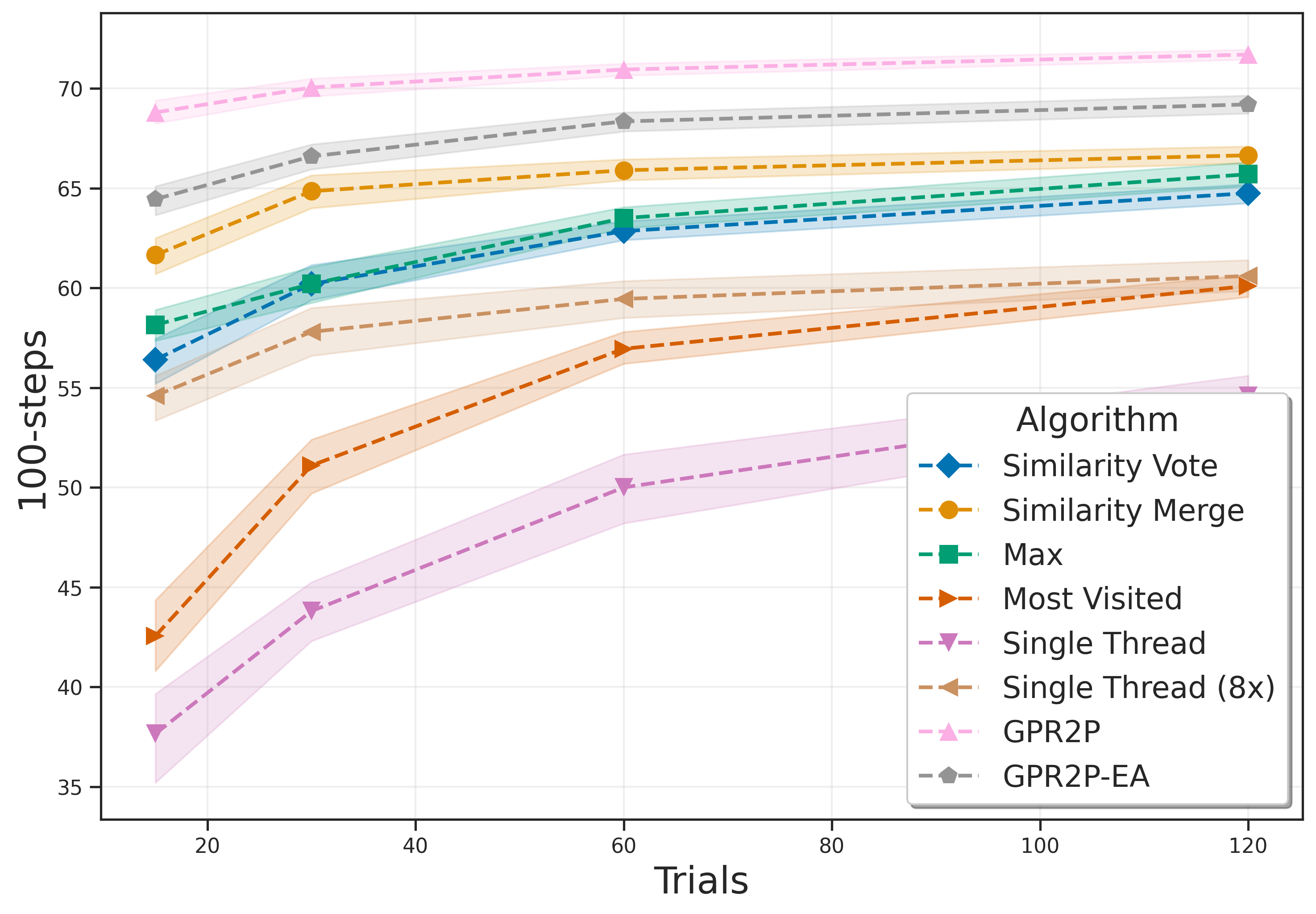}
        \caption{Narrow Corridor}
        \label{result:Narrow Corridor}
    \end{subfigure}
    \caption{Results obtained by root-parallel MCTS aggregation strategies across all environments. GPR2P performs best overall, followed by Similarity Merge. As expected, the differences diminish as the number of trials increases.}
    \label{result:overall}
\end{figure*}

\begin{figure*}[t]
\centering
    \begin{subfigure}{0.3\textwidth}
        \includegraphics[width=\linewidth, keepaspectratio]{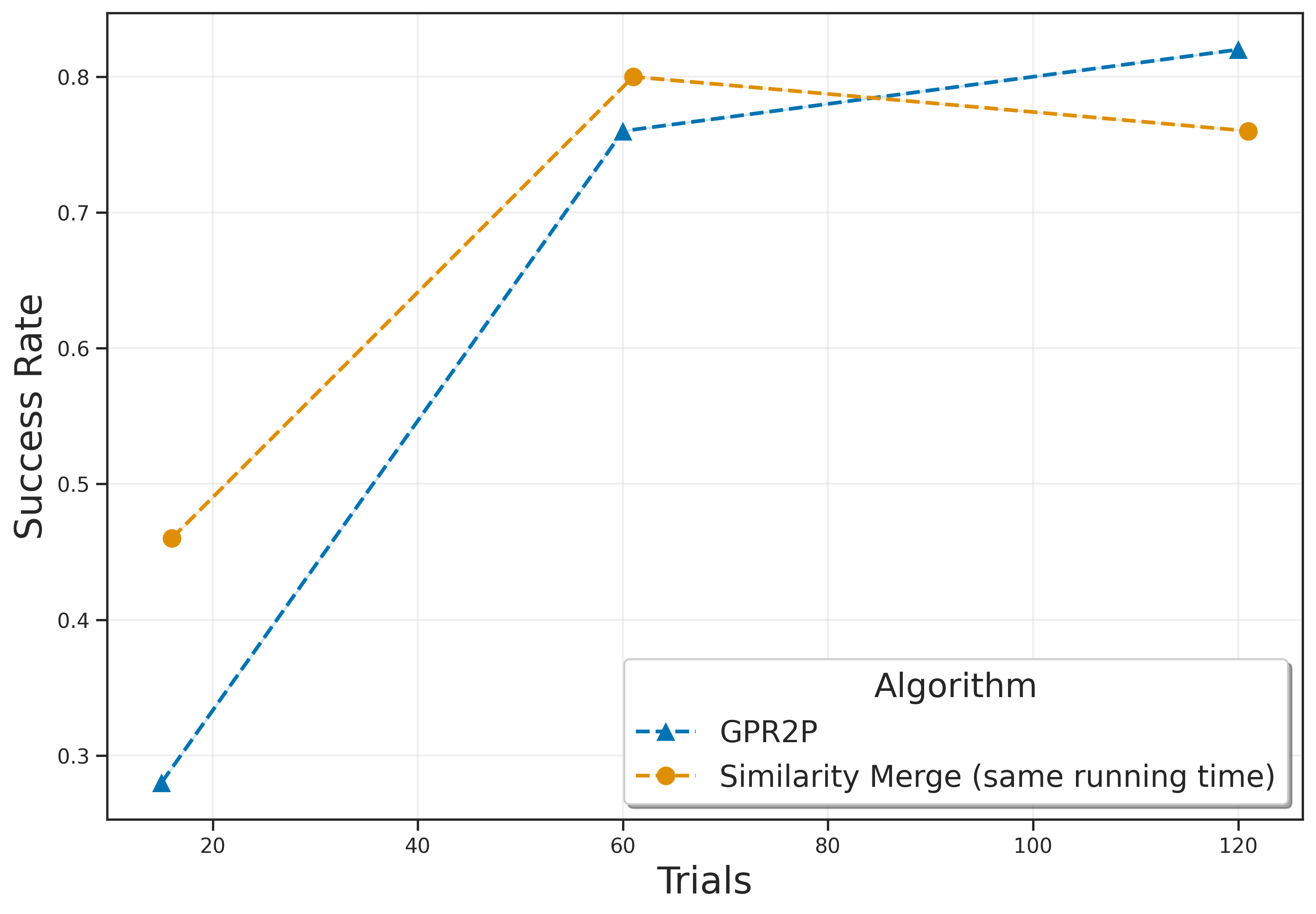}
        \caption{Lunar Lander}
        \label{extra:Lunar Lander}
    \end{subfigure}
    \begin{subfigure}{0.3\textwidth}
        \includegraphics[width=\linewidth, keepaspectratio]{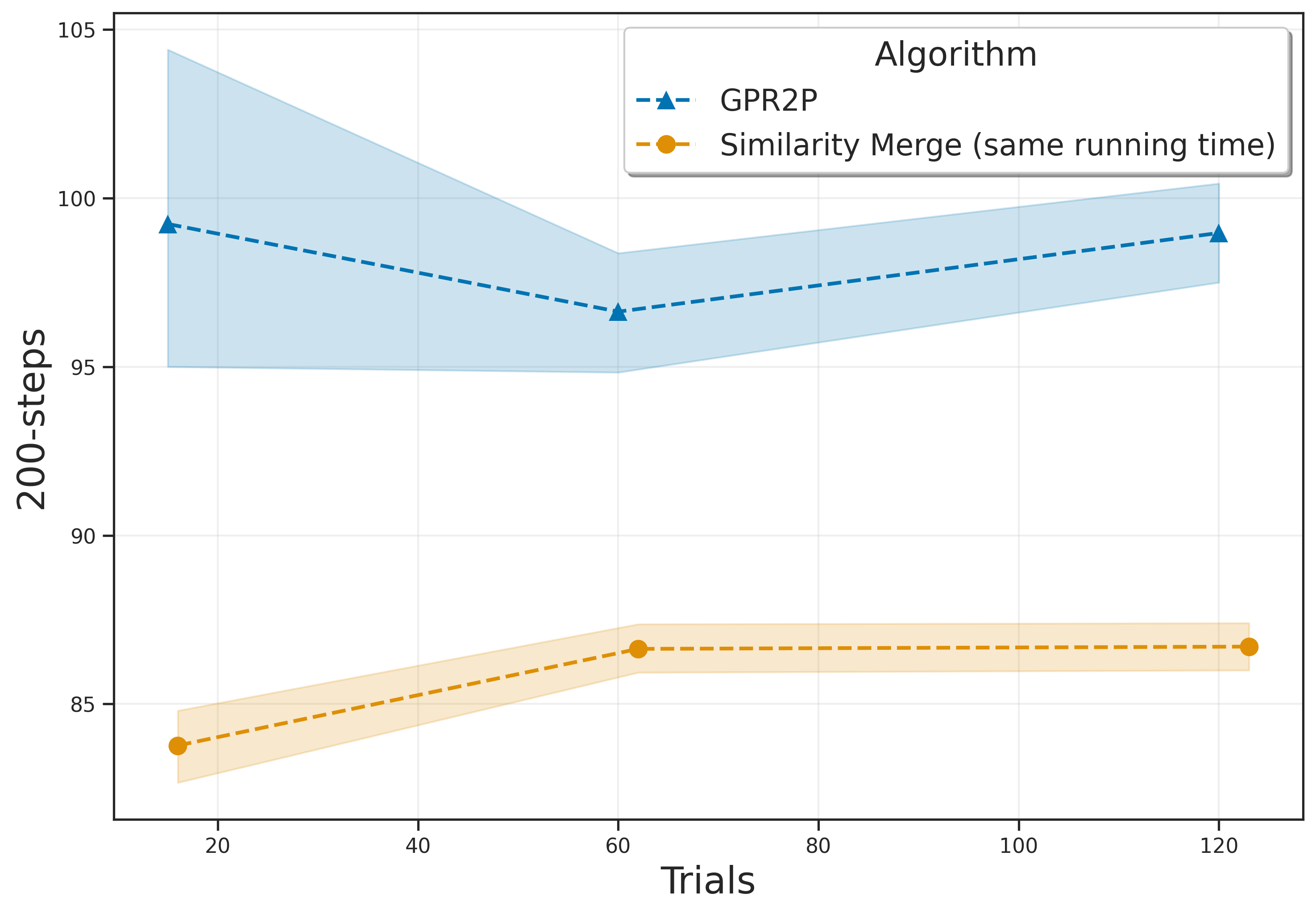}
        \caption{Mountain Car}
        \label{extra:Mountain Car}
    \end{subfigure}
    \begin{subfigure}{0.3\textwidth}
        \includegraphics[width=\linewidth, keepaspectratio]{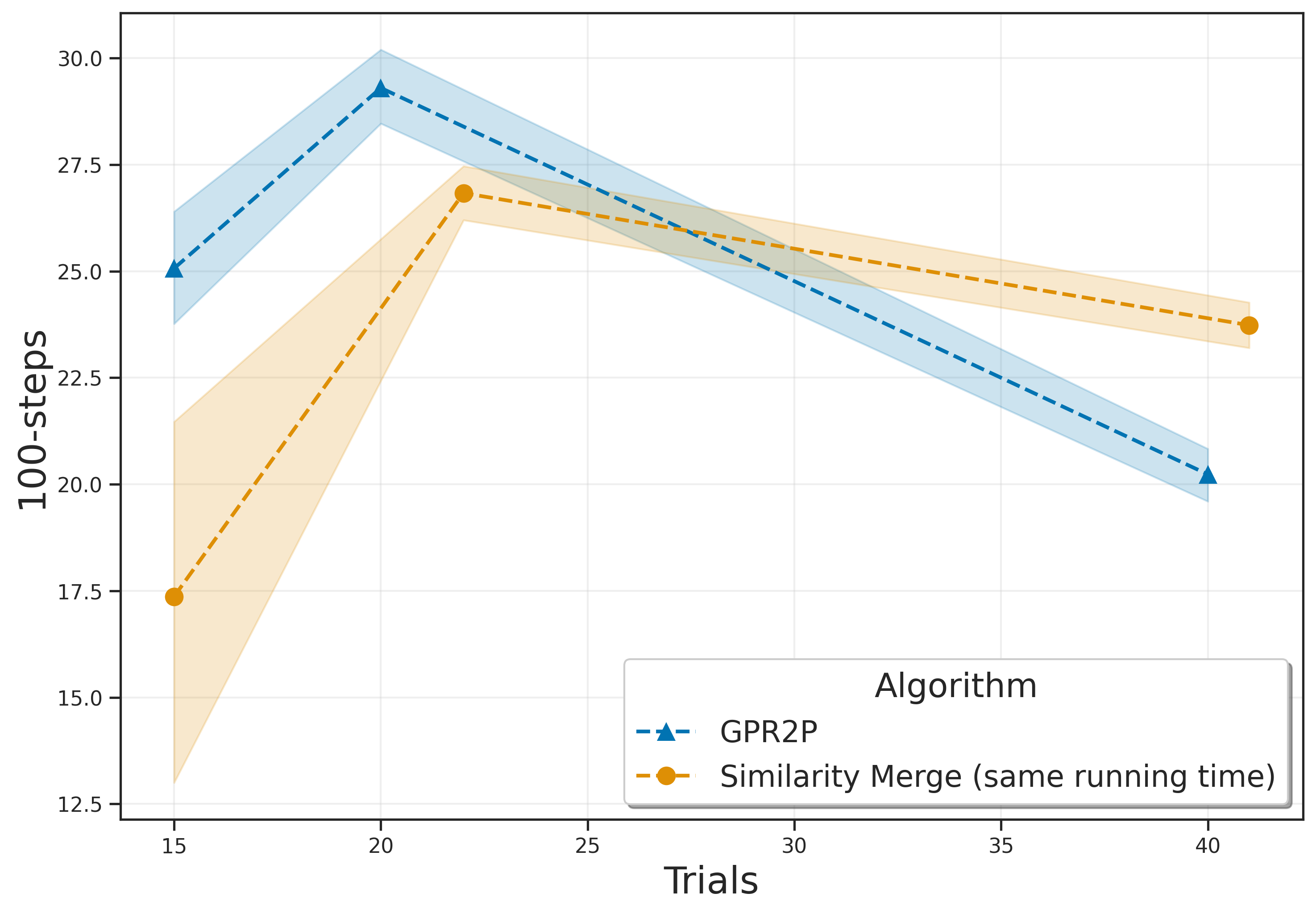}
        \caption{Pendulum}
        \label{extra:Pendulum}
    \end{subfigure}

    \begin{subfigure}{0.3\textwidth}
        \includegraphics[width=\linewidth, keepaspectratio]{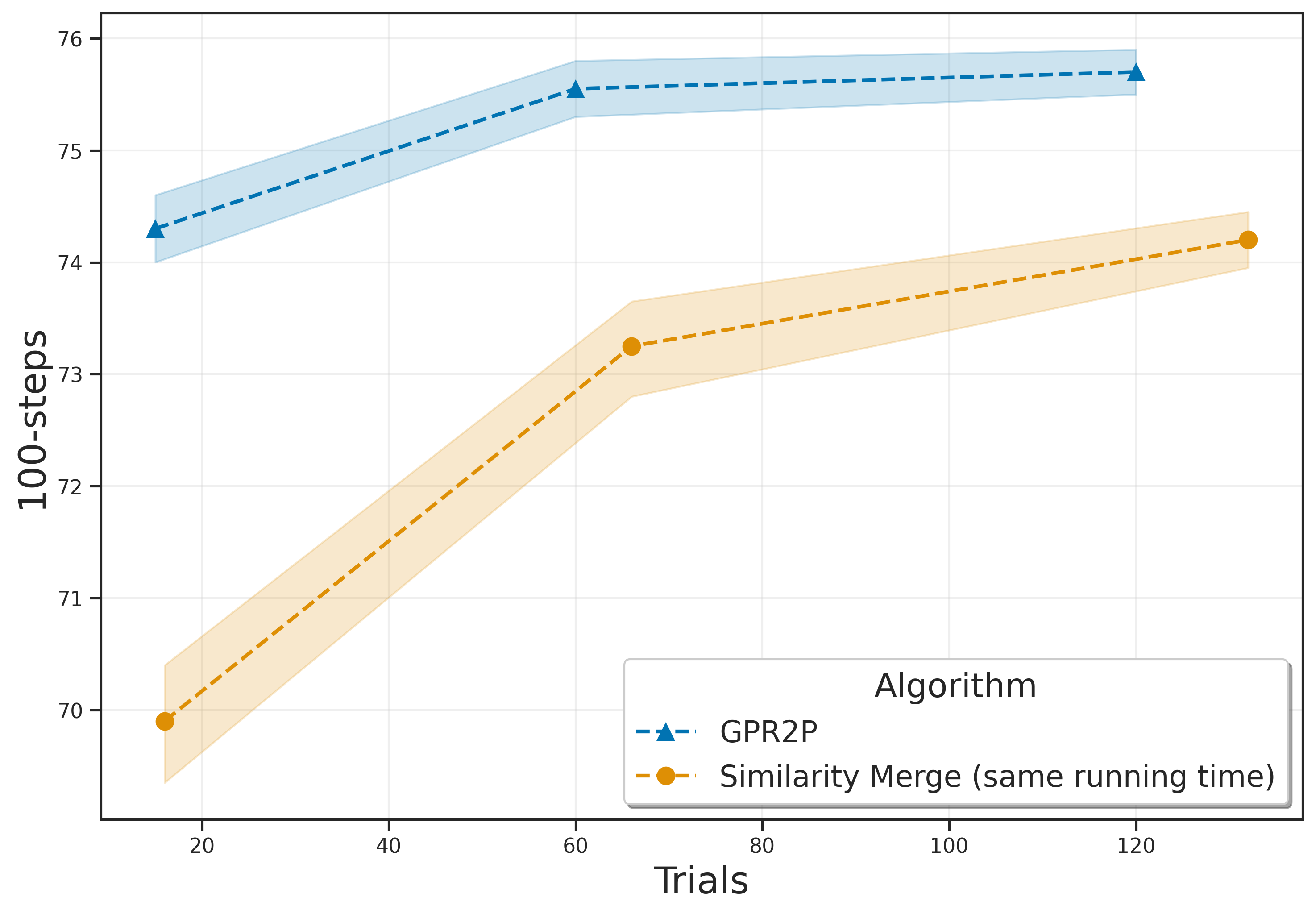}
        \caption{Random Teleporter}
        \label{extra:Random Teleporter}
    \end{subfigure}
    \begin{subfigure}{0.3\textwidth}
        \includegraphics[width=\linewidth, keepaspectratio]{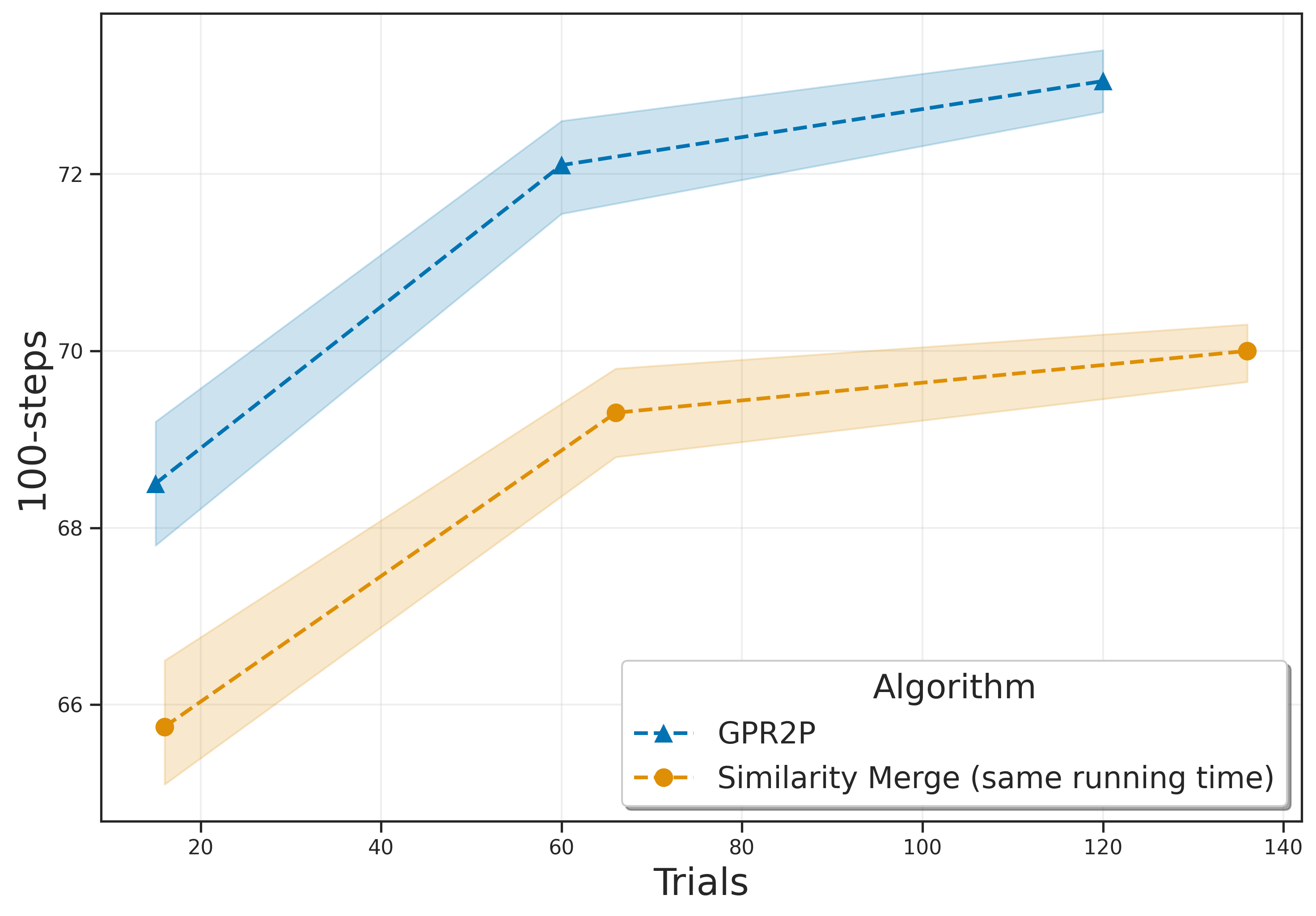}
        \caption{Wide Corridor}
        \label{extra:Wide Corridor}
    \end{subfigure}
    \begin{subfigure}{0.3\textwidth}
        \includegraphics[width=\linewidth, keepaspectratio]{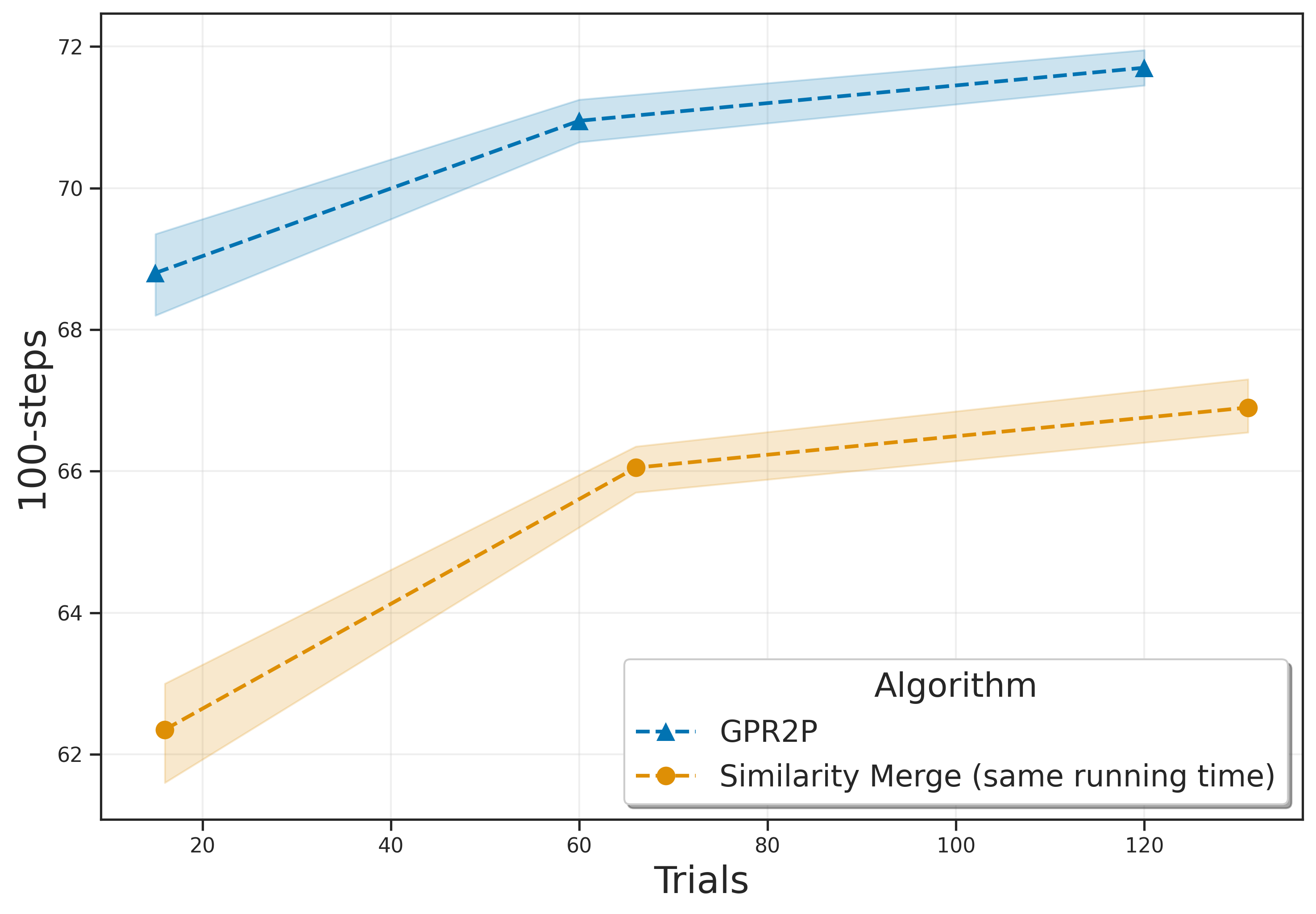}
        \caption{Narrow Corridor}
        \label{extra:Narrow Corridor}
    \end{subfigure}
    \caption{Performance comparison of GPR2P versus Similarity Merge in which the GPR2P inference time is used to run additional trials. This does not lead a significant change in the results, highlighting the value of the GPR2P aggregation.}
    \label{extra:overall}
\end{figure*}

\begin{table*}[t]
\centering
\small
\resizebox{0.95\textwidth}{!}{
\begin{tabular}{lccc cccc}
\toprule
& \multicolumn{3}{c}{\textbf{Lunar Lander}} && \multicolumn{3}{c}{\textbf{Mountain Car}} \\
Algorithm & \textit{Low \# Trials} & \textit{Mid \# Trials} & \textit{High \# Trials} && \textit{Low \# Trials} & \textit{Mid \# Trials} & \textit{High \# Trials} \\
\midrule
Similarity Vote     & 0.015 / 21.0 & 0.015 / 41.3 & 0.015 / 73.6 && 0.037 / 58.1 & 0.023 / 63.1 & 0.022 / 85.8 \\
Similarity Merge    & 0.065 / 29.9 & 0.091 / 64.8 & 0.105 / 102.4 && 0.071 / 40.4 & 0.064 / 53.1 & 0.072 / 74.6 \\
Max                   & 0.007 / 24.7 & 0.007 / 55.2 & 0.007 / 90.1 && 0.009 / 48.6 & 0.008 / 60.3 & 0.009 / 81.7 \\
Most Visited        & 0.007 / 21.7 & 0.012 / 49.5 & 0.011 / 92.7 && 0.018 / 72.0 & 0.016 / 68.9 & 0.015 / 84.2 \\
Single Thread  & 0.002 / 20.2 & 0.002 / 40.8 & 0.003 / 83.3 && 0.003  / 48.8 & 0.003 / 86.5 & 0.003 / 95.7 \\
Single Thread (8x) & 0.002 / 83.3 & 0.002 / 139.0 & 0.002 / 276.8 && 0.003 / 95.7 & 0.001 / 96.1 & 0.001 / 173.1 \\
GPR2P    & 0.860 / 31.8 & 0.873 / 74.3 & 0.827 / 110.6 && 2.501 / 71.0 & 2.012 / 59.9 & 1.990 / 75.4 \\
GPR2P-EA & 1.877 / 51.9 & 0.838 / 80.0 & 0.825 / 116.9 && 2.755 / 50.4 & 1.807 / 59.1 & 2.146 / 77.1 \\

\midrule
& \multicolumn{3}{c}{\textbf{Pendulum}} && \multicolumn{3}{c}{\textbf{Random Teleporter}} \\
Algorithm & \textit{Low \# Trials} & \textit{Mid \# Trials} & \textit{High \# Trials} && \textit{Low \# Trials} & \textit{Mid \# Trials} & \textit{High \# Trials} \\
\midrule
Similarity Vote     & 0.031 / 64.0 & 0.016 / 32.3 & 0.016 / 35.8 && 0.008 / 6.9 & 0.006 / 6.7 & 0.006 / 7.9 \\
Similarity Merge    & 0.092 / 56.6 & 0.069 / 25.6 & 0.069 / 26.9 && 0.022 / 5.7 & 0.022 / 6.0 & 0.026 / 6.9 \\
Max                  & 0.015 / 72.4 & 0.006 / 33.4 & 0.006 / 32.3 && 0.002 / 6.0 & 0.002 / 6.2 & 0.002 / 7.2 \\
Most Visited        & 0.025 / 75.5 & 0.012 / 47.3 & 0.008 / 31.9 && 0.003 / 7.6 & 0.003 / 7.1 & 0.003 / 8.2 \\
Single Thread  & 0.001 / 46.4 & 0.002 / 42.1 & 0.002 / 37.8 && 0.001 / 8.9 & 0.001 / 7.8 & 0.001 / 8.7 \\
Single Thread (8x) & 0.001 / 15.0 & 0.001 / 17.5 & 0.001 / 23.4 && 0.001 / 8.7 & 0.001 / 7.5 & 0.001 / 12.1 \\
GPR2P    & 0.362 / 25.6 & 1.560 / 31.8 & 0.466 / 29.1 && 0.229 / 5.9 & 1.034 / 9.8 & 1.166 / 11.2 \\
GPR2P-EA & 0.365 / 25.3 & 1.780 / 34.4 & 0.498 / 29.8 && 0.250 / 10.3 & 0.856 / 13.2 & 1.051 / 14.4 \\ 

\midrule
& \multicolumn{3}{c}{\textbf{Wide Corridor}} && \multicolumn{3}{c}{\textbf{Narrow Corridor}} \\
Algorithm & \textit{Low \# Trials} & \textit{Mid \# Trials} & \textit{High \# Trials} && \textit{Low \# Trials} & \textit{Mid \# Trials} & \textit{High \# Trials} \\
\midrule
Similarity Vote     & 0.010 / 8.6 & 0.007 / 8.3 & 0.008 / 10.0 && 0.013 / 14.3 & 0.009 / 10.4 & 0.008 / 10.5 \\ 
Similarity Merge    & 0.027 / 7.0 & 0.027 / 7.6 & 0.031 / 8.6 && 0.029 / 7.5 & 0.029 / 8.2 & 0.035 / 9.6 \\
Max                 & 0.003 / 7.9 & 0.002 / 8.1 & 0.002 / 9.2 && 0.003 / 8.2 & 0.002 / 8.6 & 0.002 / 9.8 \\
Most Visited        & 0.004 / 9.5 & 0.004 / 9.3 & 0.004 / 10.4 && 0.005 / 11.1 & 0.004 / 10.4 & 0.005 / 11.3 \\
Single Thread  & 0.002 / 10.3 & 0.001 / 10.6 & 0.001 / 11.3 && 0.002 / 12.0 & 0.001 / 11.8 & 0.001 / 13.4 \\
Single Thread (8x) & 0.001 / 5.3 & 0.001 / 9.3 & 0.001 / 15.1 && 0.001 / 13.4 & 0.001 / 10.4 & 0.001 / 17.2 \\
GPR2P    & 0.275 / 7.1 & 1.181 / 11.7 & 1.852 / 14.1 && 0.280 / 7.1  & 1.330 / 12.3 & 1.195 / 13.5 \\
GPR2P-EA & 0.292 / 12.3 & 1.179 / 15.9 & 1.447 / 17.6 && 0.291 / 12.1 & 1.188 / 16.3 & 1.280 / 17.7 \\

\bottomrule
\end{tabular}
}
\caption{\chl{Mean inference time required to run the root-parallel aggregation method and infer the best action (left value) and the total MCTS runtime including all aggregation steps (right value). Values are reported in seconds and are measured per episode across the six considered environments. Different MCTS trial budgets (Low / Mid / High) were investigated, which correspond to the three trial budgets (x-axis) used for GPR2P in Figure~\ref{extra:overall}. Overall, GPR2P requires a modest increase in inference time relative to the total runtime.
}}
\label{tab:Time comparison table}
\end{table*}

\begin{table*}[!t]
\centering
\resizebox{0.95\textwidth}{!}{
\begin{tabular}{|l|c|c|c|c|c|c||c|}
\hline
\textbf{Algorithm} & \textbf{LL} & \textbf{MC} & \textbf{PE} & \textbf{RT} & \textbf{WC} & \textbf{NC} & \textbf{Overall Average MRR} \\
\hline
Similarity Vote & 0.4250 & 0.1667 & 0.1756 & 0.2125 & 0.2250 & 0.2125 & 0.2362 \\
\hline
Similarity Merge & \textbf{0.8125} & 0.3333 & 0.6667 & 0.4167 & 0.3333 & 0.3333 & 0.4826 \\
\hline
Max & 0.2292 & 0.2375 & 0.1917 & 0.2375 & 0.2250 & 0.2500 & 0.2285 \\
\hline
Most Visited & 0.1488 & 0.1384 & 0.1795 & 0.1429 & 0.1429 & 0.1429 & 0.1492 \\
\hline
Single Thread & 0.1399 & 0.1295 & 0.1339 & 0.1250 & 0.1250 & 0.1250 & 0.1297 \\
\hline
Single Thread (8x) & 0.1750 & 0.2167 & 0.2167 & 0.1667 & 0.1667 & 0.1667 & 0.1847 \\
\hline
GPR2P & 0.7500 & \textbf{1.0000} & 0.6667 & \textbf{1.0000} & \textbf{1.0000} & \textbf{1.0000} & \textbf{0.9028} \\
\hline
GPR2P-EA & 0.6667 & \textbf{1.0000} & \textbf{0.7500} & 0.4167 & 0.5000 & 0.5000 & 0.6389 \\
\hline
\end{tabular}
}
\caption{Mean Reciprocal Rank (MRR) of aggregation methods averaged across all experiments. \textit{Abbreviations: LL (Lunar Lander), MC (Mountain Car), PE (Pendulum), RT (Random Teleporter), WC (Wide Corridor), NC (Narrow Corridor).}}
\label{tab:rank}
\end{table*}

\subsection{Evaluation Procedure}

We conduct experiments in six environments: Lunar Lander, Mountain Car, Pendulum, Random Teleporter, Wide Corridor, and Narrow Corridor. These environments, illustrated in Figure~\ref{fig:envs}, are control tasks that are part of Gymnasium~\citep{brockman2016openai,towers2024gymnasium} or self-designed continuous action tasks. Descriptions of the environments are provided in the Supplementary Material. The rationale for developing the self-designed tasks is that transitions are deterministic in these standard environments, whereas many practical planning problems exhibit stochastic transitions. For example, in search-and-rescue, a robot may be assigned to search a particular area but lack communication while the task is being carried out. Similarly, in mission planning for underwater gliders, actions are decided while the robot is at the surface and can communicate via satellite but is unreachable while underwater. We aim to demonstrate that our technique generalises to both environments with deterministic and stochastic transitions. Progressive Widening is thus used in UCT in the Gymnasium environments, while Double Progressive Widening is required in the self-designed ones to account for stochastic transitions.

Experiments are conducted using 8 parallel threads, yielding 8 trees at each step when running root-parallel MCTS. Alongside GPR2P, we evaluate GPR2P with Existing Actions (GPR2P-EA), a variant which selects the sampled action closest in Euclidean distance to the estimated optimal action. This enables judging the benefits of allowing GPR2P to output actions not tried in the environment. GPR2P and GPR2P-EA are compared with the prior methods described in the Methods section. To assess the effect of parallelization, we also include a Single-Thread MCTS baseline. Furthermore, we evaluate a Single-Thread MCTS (8x) variant with eight times the simulation budget, matching the total number of trials used by eight parallel threads. This allows us to isolate the contribution of the aggregation strategy from the mere increase in computational resources. Details about the hyperparameter tuning procedure and values used are supplied in the Supplementary Material.

Since the search tree information in each root-parallel thread serves as the source of raw data and directly influences the quality of the samples, we limit tree growth by fixing the number of trials. For each trial setting and method, we average results over $20$ to $50$ random seeds depending on the variability of the environment. The number of trials ranges from $15$ to $120$, as the main improvements are observed within this range, while changes beyond $120$ trials are relatively small.

To compare the performance of different methods, we adopt two evaluation metrics. The \textit{Steps} metric is used for all environments except Lunar Lander. It counts the number of steps required to reach the goal. In result plots, we show a constant minus the step count, so that higher is better across all environments. \textit{Success rate} is defined as the average percentage of trajectories that successfully complete the task (higher is better). In the Lunar Lander task, step count is not a meaningful metric, as the agent can crash rapidly, terminating the episode early. A smooth and safe landing often requires taking additional steps for fine adjustments. We note that the success rate is computed across all runs and does not enable calculating a confidence interval for Lunar Lander.

To summarize the overall performance of the algorithms across trial counts in all six tasks, we analyze the Mean Reciprocal Rank (MRR) with respect to the two evaluation metrics. As the metrics and their scales differ between the different tasks, simply averaging their values is inappropriate, while a ranking-based metric is well-suited. MRR yields a single number for each aggregation strategy that summarizes its performance across the experiments. In information retrieval, the Reciprocal Rank (RR) measures the position at which the first relevant item is retrieved, where the RR equals $1/r$ if the item appears at rank $r$ \citep{craswell2016mean}. Each of the $24$ task and trial number combinations can be treated as individual ``queries". For each task, we first compute the RR of every algorithm at each trial case and then calculate the Mean Reciprocal Rank (MRR) for the task by averaging the RR values over all trial cases. Finally, we compute the overall MRR across all six tasks to obtain the final ranking of the algorithms.

\subsection{Results}
The average performance of the methods is reported in Figure~\ref{result:overall}. When considering the results across all six environments, we observe that the rankings are largely consistent, with only minor differences. GPR2P consistently achieves the best performance, while Similarity Merge is the strongest among the existing baseline methods.

Comparing GPR2P with GPR2P-EA allows us to clearly separate the sources of improvement, namely the principled return estimation provided by Gaussian Process Regression and the benefit of explicitly selecting previously unsampled actions. In 4 out of 6 environments, removing the unsampled-action selection mechanism directly degrades performance. Sampled actions provide a coarse global picture of where the optimal action may lie, since GP integrates all observed returns to form a holistic evaluation of the action space. The Corridor environments are specifically designed to illustrate this effect. In the Narrow Corridor setting, where the rewarding region is extremely limited, performance gains from increasing samples quickly reach a bottleneck, whereas this effect is less pronounced in the Wide Corridor case. Consequently, most existing methods exhibit a clear convergence tendency in the 120-trial scenario. In contrast, this bottleneck has much less impact on the GP-based methods.

In contrast to the consistent advantage seen in most environments, we observe a relative performance drop of the GPR2P methods in the Pendulum task as the number of trials increases. While GPR2P outperforms the other methods when the number of trials is small, its advantage diminishes once the trial count exceeds 30, at which point its performance becomes comparable to that of Similarity Merge. By examining the trajectories, this later-stage shift can be attributed to the agent discovering a gravity-assisted strategy that involves an additional swing. Although this strategy requires more steps than the direct path, it is more action-efficient and leads to better overall resource utilization.

Across all tasks, we observe that Similarity Vote and Max exhibit highly similar performance, with only minor differences. This behavior arises because the voting mechanism primarily depends on the values of the sampled actions, and the only parameter influencing the system is the one measuring the distance between actions. When this parameter is optimized via grid search, it tends to select values that consider only very close actions as similar. Consequently, the two methods behave almost identically, which explains their comparable outcomes. We also observe that Similarity Merge consistently outperforms Similarity Vote across all six tasks, which differs from the findings of \citet{kurzer2020parallelization}. One possible reason is that their evaluation was conducted on a single environment, which may limit the generality of the comparison. In contrast, in our experiments across six environments, Similarity Merge consistently shows the best performance. 

Most Visited does not achieve strong performance and only slightly outperforms Single Thread MCTS. In root-parallel MCTS, since there is no merging of visit counts across trees, Most Visited effectively treats each parallel tree independently, and comparing action visit counts across different trees is not intuitively convincing because the number of visits for each action is highly dependent on the specific trajectory of tree construction. Single Thread (8x), despite using significantly longer wall clock time, only converges faster but does not achieve better final performance compared to Single Thread. This result highlights the importance of an effective aggregation strategy and demonstrates that the advantage of parallel methods does not solely arise from increased computational resources provided by parallelization.

While the GPR2P aggregation provides additional information, it also incurs extra computational cost. A natural question is whether better performance would be obtained by using the time required for GPR2P inference to run additional trials instead. Table~\ref{tab:Time comparison table} reports the average inference time and total runtime per episode for all tasks and algorithms. We observe that the inference time of GPR2P is higher than those of the other methods. However, this overhead is relatively minor compared to the overall time spent on data collection. To assess the impact of the additional inference time, we test GPR2P against Similarity Merge with the corresponding time allocated to run additional trials. As shown in Figure~\ref{extra:overall}, the advantage of GPR2P relative to Similarity Merge is preserved.

The MRR obtained by the methods is given in Table~\ref{tab:rank}, providing a comprehensive overview of performance across all environments and trial regimes. It is evident that GPR2P consistently outperforms all other approaches. While its computational cost is slightly higher than the alternatives, its superior performance makes it the most effective aggregation algorithm for root-parallel MCTS.

\section{Conclusion and Future Work}
In this paper, we investigate aggregation strategies for root-parallel MCTS in environments with continuous action spaces. We propose a new aggregation method, GPR2P, which leverages Gaussian Process Regression to estimate returns over the entire action space. By doing so, it overcomes the key limitations of the state-of-the-art approaches proposed by~\citet{kurzer2020parallelization}, which aggregate information from actions in an ad-hoc way and restrict action selection to those already sampled in the tree. We conduct a comprehensive evaluation across six environments with both deterministic and stochastic transitions, comparing GPR2P against several aggregation strategies. The results demonstrate that our method consistently outperforms the alternatives.

A disadvantage of GPR2P is that it requires additional parameters (the threshold $\tau$ and GPR parameters). However, Similarity Vote and Similarity Merge also require parameter tuning. If this can be performed offline prior to deploying the system for online planning, the advantage of GPR2P is well justified. \chl{Another limitation is the inherent assumption that similar actions tend to have similar values, which may be less suitable for highly non-linear or discontinuous value landscapes.} Furthermore, there is also potential for using the GP models to also guide action selection \textit{within} each individual MCTS thread. This research direction was explored by~\cite{yee2016monte}, and we consider that further improvements can be obtained by combining the two methods.

\bibliography{main}

@inproceedings{kocsis2006bandit,
  title={Bandit based {Monte-Carlo} planning},
  author={Kocsis, Levente and Szepesv{\'a}ri, Csaba},
  booktitle={European Conference on Machine Learning},
  year={2006},
}

@inproceedings{cazenave2007parallelization,
  title={On the parallelization of {UCT}},
  author={Cazenave, Tristan and Jouandeau, Nicolas},
  booktitle={Computer Games Workshop},
  year={2007}
}

@article{chaslot2008progressive,
  title={Progressive strategies for {Monte-Carlo} tree search},
  author={Chaslot, Guillaume M. J. B. and Winands, Mark H. M. and Herik, H. Jaap van den and Uiterwijk, Jos W. H. M. and Bouzy, Bruno},
  journal={New Mathematics and Natural Computation},
  volume={4},
  number={03},
  pages={343--357},
  year={2008},
  publisher={World Scientific}
}

@inproceedings{coulom2006efficient,
  title={Efficient selectivity and backup operators in {Monte-Carlo} tree search},
  author={Coulom, R{\'e}mi},
  booktitle={International Conference on Computers and Games},
  year={2006},
}

@article{kurzer2020parallelization,
  title={Parallelization of {Monte Carlo} tree search in continuous domains},
  author={Kurzer, Karl and H{\"o}rtnagl, Christoph and Z{\"o}llner, J Marius},
  journal={arXiv preprint arXiv:2003.13741},
  year={2020}
}

@inproceedings{williams1995gaussian,
  title={Gaussian processes for regression},
  author={Williams, Christopher K. I. and Rasmussen, Carl Edward},
  journal={Advances in Neural Information Processing Systems},
  volume={8},
  year={1995}
}

@book{rasmussen2006gaussian,
  title={Gaussian Processes for Machine Learning},
  author={Rasmussen, Carl Edward and Williams, Christopher K. I.},
  year={2006},
  publisher={MIT Press}
}

@article{silver2016mastering,
  title={Mastering the game of {Go} with deep neural networks and tree search},
  author={Silver, David and Huang, Aja and Maddison, Chris J and Guez, Arthur and Sifre, Laurent and van den Driessche, George and Schrittwieser, Julian and Antonoglou, Ioannis and Panneershelvam, Veda and Lanctot, Marc and others},
  journal={Nature},
  volume={529},
  number={7587},
  pages={484--489},
  year={2016},
  publisher={Nature Publishing Group}
}

@inproceedings{chaslot2008parallel,
  title={Parallel {Monte-Carlo} tree search},
  author={Chaslot, Guillaume M. J. B. and Winands, Mark H. M. and van den Herik, H. Jaap},
  booktitle={International Conference on Computers and Games},
  year={2008},
}

@article{soejima2010evaluating,
  title={Evaluating root parallelization in {Go}},
  author={Soejima, Yusuke and Kishimoto, Akihiro and Watanabe, Osamu},
  journal={IEEE Transactions on Computational Intelligence and AI in Games},
  volume={2},
  number={4},
  pages={278--287},
  year={2010},
  publisher={IEEE}
}

@article{brockman2016openai,
  title={{OpenAI Gym}},
  author={Brockman, Greg and Cheung, Vicki and Pettersson, Ludwig and Schneider, Jonas and Schulman, John and Tang, Jie and Zaremba, Wojciech},
  journal={arXiv preprint arXiv:1606.01540},
  year={2016}
}

@inproceedings{towers2024gymnasium,
  title={Gymnasium: A standard interface for reinforcement learning environments},
  author={Towers, Mark and Kwiatkowski, Ariel and Terry, Jordan and Balis, John U. and De Cola, Gianluca and Deleu, Tristan and Goul{\~a}o, Manuel and Kallinteris, Andreas and Krimmel, Markus and KG, Arjun and others},
  booktitle={Advances in Neural Processing Systems},
  volume={38},
  year={2025}
}

@inproceedings{couetoux2011continuous,
  title={Continuous upper confidence trees},
  author={Cou{\"e}toux, Adrien and Hoock, Jean-Baptiste and Sokolovska, Nataliya and Teytaud, Olivier and Bonnard, Nicolas},
  booktitle={International Conference on Learning and Intelligent Optimization},
  year={2011},
}

@incollection{craswell2016mean,
  title={Mean reciprocal rank},
  author={Craswell, Nick},
  booktitle={Encyclopedia of Database Systems},
  year={2016},
  publisher={Springer}
}

@article{bellman1957markovian,
  title={A {Markovian} decision process},
  author={Bellman, Richard},
  journal={Journal of Mathematics and Mechanics},
  pages={679--684},
  year={1957},
  publisher={JSTOR}
}

@inproceedings{silver2010monte,
  title={{Monte-Carlo} planning in large {POMDPs}},
  author={Silver, David and Veness, Joel},
  booktitle={Advances in Neural Information Processing Systems},
  volume={23},
  year={2010}
}

@article{browne2012survey,
  title={A survey of {Monte Carlo} tree search methods},
  author={Browne, Cameron B. and Powley, Edward and Whitehouse, Daniel and Lucas, Simon M. and Cowling, Peter I. and Rohlfshagen, Philipp and Tavener, Stephen and Perez, Diego and Samothrakis, Spyridon and Colton, Simon},
  journal={IEEE Transactions on Computational Intelligence and AI in Games},
  volume={4},
  number={1},
  pages={1--43},
  year={2012},
  publisher={IEEE}
}

@article{steinmetz2020more,
  title={More trees or larger trees: parallelizing {Monte Carlo} tree search},
  author={Steinmetz, Erik and Gini, Maria},
  journal={IEEE Transactions on Games},
  volume={13},
  number={3},
  pages={315--320},
  year={2020},
  publisher={IEEE}
}

@article{scholkopf1997comparing,
  title={Comparing support vector machines with {Gaussian} kernels to radial basis function classifiers},
  author={Scholkopf, Bernhard and Sung, Kah-Kay and Burges, Christopher J. C. and Girosi, Federico and Niyogi, Partha and Poggio, Tomaso and Vapnik, Vladimir},
  journal={IEEE Transactions on Signal Processing},
  volume={45},
  number={11},
  pages={2758--2765},
  year={1997},
  publisher={IEEE}
}

@article{mackay1998introduction,
  title={Introduction to {Gaussian} processes},
  author={MacKay, David J.C.},
  journal={NATO ASI Series F Computer and Systems Sciences},
  volume={168},
  pages={133--166},
  year={1998},
  publisher={Springer Verlag}
}

@inproceedings{yee2016monte,
  title={{Monte Carlo} Tree Search in Continuous Action Spaces with Execution Uncertainty},
  author={Yee, Timothy and Lis{\`y}, Viliam and Bowling, Michael H. and Kambhampati, S.},
  booktitle={International Joint Conference on Artificial Intelligence},
  year={2016}
}

@inproceedings{enzenberger2009lock,
  title={A lock-free multithreaded {Monte-Carlo} tree search algorithm},
  author={Enzenberger, Markus and M{\"u}ller, Martin},
  booktitle={Advances in Computer Games},
  year={2009},
}

@inproceedings{gelly2008parallelization,
  title={On the parallelization of {Monte-Carlo} planning},
  author={Gelly, Sylvain and Hoock, Jean-Baptiste and Rimmel, Arpad and Teytaud, Olivier and Kalemkarian, Yann},
  booktitle={ICINCO},
  year={2008}
}

@inproceedings{cazenave2015generalized,
  title={Generalized rapid action value estimation},
  author={Cazenave, Tristan},
  booktitle={24th International Conference on Artificial Intelligence},
  year={2015}
}

@inproceedings{sokota2021monte,
  title={{Monte Carlo} tree search with iteratively refining state abstractions},
  author={Sokota, Samuel and Ho, Caleb Y. and Ahmad, Zaheen and Kolter, J. Zico},
  booktitle={Advances in Neural Information Processing Systems},
  volume={34},
  year={2021}
}

@article{coulom2007computing,
  title={Computing {Elo} ratings of move patterns in the game of {Go}},
  author={Coulom, R{\'e}mi},
  journal={ICGA Journal},
  volume={30},
  number={4},
  pages={198--208},
  year={2007},
  publisher={SAGE}
}

@article{pedregosa2011scikit,
  title={Scikit-learn: Machine learning in {Python}},
  author={Pedregosa, Fabian and Varoquaux, Ga{\"e}l and Gramfort, Alexandre and Michel, Vincent and Thirion, Bertrand and Grisel, Olivier and Blondel, Mathieu and Prettenhofer, Peter and Weiss, Ron and Dubourg, Vincent and others},
  journal={Journal of Machine Learning Research (JMLR)},
  volume={12},
  pages={2825--2830},
  year={2011},
}
\bibliographystyle{rlj}

\beginSupplementaryMaterials

\section*{Environment Descriptions}

\textit{Lunar Lander} is a task in which the agent must control the bottom and side engines of a spacecraft to ensure a safe and smooth landing on the pad. The agent starts in a random position with a random orientation and velocity. Both the state space and the action space are 2-dimensional. Since the relative orientation of the engines and the lander body is fixed, the agent controls the thrust magnitude to balance fuel efficiency with a successful landing, making the action space continuous.

\textit{Mountain Car} is a one-dimensional control problem. The agent applies left or right forces to move the car forward or backward. Since the applied force is insufficient to drive the car directly over the hill, the agent must leverage the gravitational potential and the terrain to reach the goal.

\textit{Pendulum} is a task in which the agent applies clockwise or counterclockwise torque to control a pendulum. The goal is to learn a control policy that applies appropriate forces to stabilize the pendulum in the upright position with zero angular velocity. 

\textit{Random Teleporter} is a path-finding problem with a 2-dimensional continuous action space and state space. The task requires the agent to reach the goal in as few steps as possible. Transitions are stochastic and involve the addition of randomly sampled noise to the direction and magnitude of the selected action. 

\textit{Wide Corridor} and \textit{Narrow Corridor} are variants of the Random Teleporter task. The state and action spaces remain the same; the difference is in the transition function, which applies an external force determined by the position. The magnitude and direction of this force can either assist or hinder movement toward the goal, so the agent must identify a good path in order to make effective use of the external force. Specifically, only a portion of the space, shaped like a corridor, is beneficial for the agent. The width of this corridor is adjusted in the two variants to simulate different levels of sampling difficulty. All other parameters remain the same as in the Random Teleporter task.

\section*{Evaluation Details}

All experiments were conducted on a machine with an AMD Ryzen 7 7845HX CPU and 16 GB RAM, running Ubuntu 22.04.5. Since the effectiveness of parallelization also depends on the construction of the search tree, we explicitly specify the hyper-parameters used in the UCT equation and the PW procedure. Although these parameters differ across tasks, they are kept consistent for all methods within each individual task. For every task, we perform a grid search by running standalone MCTS with either PW or DPW. The hyperparameter configuration that achieves the best performance is then selected.

\chl{For aggregation strategies that require specifying parameters, we carry out a parameter tuning stage.} Specifically, Similarity Vote, Similarity Merge, and GPR2P require parameter tuning. For Similarity Vote and Similarity Merge, we tune the only parameter $\phi$ which specifies the connection between exact distance and similarity, \chl{using grid search}. 

GPR2P uses a kernel composed of a signal term, modeled by an RBF kernel with variance $\alpha_f^2$ and length-scale $l$ controlling the decay of correlations with distance between actions, together with a noise term of variance $\alpha_n^2$. \chl{The kernel hyperparameters are optimized by maximizing the log marginal likelihood using the L-BFGS-B implementation in scikit-learn~\citep{pedregosa2011scikit}. We use a zero-mean GP prior on standardized action values, corresponding to their empirical mean in the original space.} In addition, GPR2P includes a threshold parameter $\tau$ used to filter sampled actions, \chl{which is selected via grid search}. GPR2P-EA uses the same parameters as GPR2P.

The parameters used in the experiments are shown in Table~\ref{tab:exp_parameters_final}. \chl{For the Mountain Car environment, we adopt a heuristic rollout policy in MCTS that applies force in the direction of the current velocity, with additional random perturbations, to encourage momentum accumulation and account for long horizons. This policy is applied consistently for all algorithms.}

\begin{table*}[t] 
    \centering
    \resizebox{\textwidth}{!}{ 
    \begin{tabular}{c | c c c c c | c | c | c c c c c c c}
        \toprule
        \multirow{2}{*}{\textbf{Environment}} & 
        \multicolumn{5}{c|}{\textbf{MCTS Parameters}} & 
        \multicolumn{1}{c|}{\textbf{Similarity Vote}} & 
        \multicolumn{1}{c|}{\textbf{Similarity Merge}} & 
        \multicolumn{7}{c}{\textbf{GPR2P}} \\ 
        
        \cmidrule(lr){2-6} \cmidrule(lr){7-7} \cmidrule(lr){8-8} \cmidrule(lr){9-15} 

        & \makecell[c]{UCT Weight \\ $C$} & \makecell[c]{PW $c$} & \makecell[c]{DPW $d$} & \makecell[c]{PW $\alpha$} & \makecell[c]{DPW $\beta$} & 
        \makecell[c]{$\phi$} & 
        \makecell[c]{$\phi$} & 
        \makecell[c]{$\alpha_f^2$} & \makecell[c]{$l$} & \makecell[c]{$\alpha_n^2$} &
        $\tau_1$ & $\tau_2$ & $\tau_3$ & $\tau_4$ \\
        \midrule

        \textbf{Lunar Lander} & 7 & 2 & \textbf{N/A} & 0.4 & \textbf{N/A} & 25 & 1.5 & 0.054 & 2.71 & 0.899 & 1 & 4 & 6 & 8 \\
        \midrule
        \textbf{Mountain Car} & 2 & 5 & \textbf{N/A} & 0.2 & \textbf{N/A} & 5 & 5 & 0.054 & 2.71 & 0.899 & 1 & 3 & 5 & 7 \\
        \midrule
        \textbf{Pendulum} & 2 & 5 & \textbf{N/A} & 0.12 & \textbf{N/A} & 25 & 5 & 0.5 & 2.5 & 0.1 & 1 & 1 & 4 & 5 \\
        \midrule
        \textbf{Random Teleporter} & 10 & 2 & 1.2 & 0.7 & 0.2 & 25 & 1 & 0.284 & 2.61 & 0.899 & 1 & 1 & 1 & 1 \\
        \midrule
        \textbf{Wide Corridor} & 10 & 2 & 1.2 & 0.7 & 0.2 & 25 & 1 & 0.284 & 2.61 & 0.899 & 1 & 1 & 1 & 1 \\
        \midrule
        \textbf{Narrow Corridor} & 10 & 2 & 1.2 & 0.7 & 0.2 & 25 & 1 & 0.284 & 2.61 & 0.899 & 1 & 1 & 1 & 1 \\
        
        \bottomrule
    \end{tabular}
    }
        \caption{Parameters used in the six experimental environments. The MCTS Parameters group includes the UCT weight $C$ and the Progressive Widening (PW) or Double Progressive Widening (DPW) parameters $c, d, \alpha, \beta$. For environments where only PW is applied, the DPW-specific parameters ($d$ and $\beta$) are denoted as N/A. The GPR2P thresholds $\tau_1, \tau_2, \tau_3, \tau_4$ correspond to four predefined trial numbers, which are set to 15, 30, 60, and 120 in all environments except Pendulum, which uses 15, 20, 30, and 40.}
        \label{tab:exp_parameters_final}

\end{table*}

\section*{Further Details on Baselines}

Pseudocode for Similarity Vote and Similarity Merge is provided in Algorithms~\ref{alg:sim_vote} and~\ref{alg:sim_merge} so that they may be compared with GPR2P using consistent notation.

\begin{algorithm}[h]
\caption{Similarity Vote~\citep{kurzer2020parallelization}}
\label{alg:sim_vote}
\begin{algorithmic}[1]  
\Require Collection of trees $\mathcal{X}$, similarity scale $\phi$
\State $\chl{\mathcal{A}^* \gets \emptyset}$
\For{$x \in \mathcal{X}$}
    \State $\mathcal{A}^* \gets \mathcal{A}^* \cup \arg\max_a Q_x(s_0,a)$
\EndFor
\For{$a_i \in \mathcal{A}^*$}
    \For{$a_j \in \mathcal{A}^*$}
        \State $K_{ij} \gets \exp\left(-\phi \lVert a_i - a_j \rVert^2\right)$
    \EndFor
    \State $v_i \gets Q(s_0, a_i)$
\EndFor
\State \Return $a_{\text{final}} \gets a_i \in \mathcal{A}^* \mid i = \arg\max_i (Kv)_i$
\end{algorithmic}
\end{algorithm}

\begin{algorithm}[h]
\caption{Similarity Merge~\citep{kurzer2020parallelization}}
\label{alg:sim_merge}
\begin{algorithmic}[1]
\Require Collection of trees $\mathcal{X}$, similarity scale $\phi$
\State $\chl{\mathcal{A}_{\text{final}} \gets \emptyset}$
\For{$x \in \mathcal{X}$, \chl{with root action set $\mathcal{A}_x$}}
    \State $\chl{\mathcal{A}_{\text{final}} \gets \mathcal{A}_{\text{final}} \cup \mathcal{A}_x}$
\EndFor
\For{$a_i \in \mathcal{A}_{\text{final}}$}
    \For{$a_j \neq a_i \in \mathcal{A}_{\text{final}}$}
        \State $K_{ij} \gets \exp\left(-\phi \lVert a_i - a_j \rVert^2\right)$
        \State $N_{\text{sim}}(s_0,a_i) \gets N(s_0,a_i) + K_{ij}N(s_0,a_j)$
        \State $Q_{\text{sim}}(s_0,a_i) \gets \frac{1}{N_{\text{sim}}}
        \Big( N(s_0,a_i)Q(s_0,a_i) + K_{ij}N(s_0,a_j)Q(s_0,a_j) \Big)$
    \EndFor
\EndFor
\State \Return $a_{\text{final}} \gets \arg\max_{a \in \mathcal{A}_{\text{final}}} Q_{\text{sim}}(s_0,a)$
\end{algorithmic}
\end{algorithm}

\end{document}